\documentclass[11pt]{article}
\usepackage[final]{acl}

\usepackage{times}
\usepackage{latexsym}
\usepackage[T1]{fontenc}
\usepackage[utf8]{inputenc}
\usepackage{microtype}
\IfFileExists{inconsolata.sty}{\usepackage{inconsolata}}{}
\usepackage{graphicx}
\usepackage{booktabs}
\usepackage{amsmath,amssymb}
\usepackage{amsthm}
\usepackage{longtable}
\usepackage{array,alphalph,placeins}
\usepackage{multirow}
\usepackage{xspace}
\usepackage{enumitem}
\usepackage{xcolor}
\usepackage{tikz}
\usetikzlibrary{positioning,arrows.meta,calc,fit,decorations.pathreplacing,backgrounds}
\usepackage{hyperref}
\usepackage[capitalize,nameinlink,noabbrev]{cleveref}

\newtheorem{proposition}{Proposition}
\theoremstyle{definition}
\newtheorem{definition}{Definition}
\crefname{observation}{Observation}{Observations}
\Crefname{observation}{Observation}{Observations}
\newcommand{\JS}{\mathrm{JS}}
\newcommand{\KL}{\mathrm{KL}}
\newcommand{\CL}{\mathrm{CL}}

\newcommand{\Own}{\mathrm{Own}}

\definecolor{clblue}{HTML}{1F6FB2}
\definecolor{jsorange}{HTML}{D9822B}
\definecolor{ownviolet}{HTML}{7A5BA6}
\definecolor{supp}{HTML}{F3D9B1}

\title{SELD: Detecting Silent Evidence Loss in Context Compression\\beyond the Answer}

\author{%
  \normalfont\normalsize\itshape Wentao Lei\textsuperscript{1}, Li Liu\textsuperscript{1,*}\\
  \normalfont\normalsize \textsuperscript{1}The Hong Kong University of Science and Technology (Guangzhou), China\\
  \texttt{avrillliu@hkust-gz.edu.cn}\\
  {\normalfont\small\textsuperscript{*}Corresponding author}%
}

\crefname{definition}{Definition}{Definitions}
\newcommand{\appref}[1]{\cref*{#1}}

\begin{document}
\maketitle

\begin{abstract}
Context compression is now routine for language models with long inputs, but
a compressed context is useful only if it keeps the evidence for the answer.
Existing checks compare the reader model's answer or its most likely words, and
thus miss a frequent failure, \emph{silent evidence loss}: key evidence is
removed while the answer stays the same, leaving the answer unsupported and
risky for other models that reuse the context. To address this challenge, we
propose SELD (Silent Evidence Loss Detector), which needs no training and
no labels. Its principle is
that when evidence is lost, a reader model's preferences among the answers it leaves
unsaid shift even if its answer does not, so comparing these preferences
before and after compression reveals losses that answer checks cannot see.
We also build SEL-Pairs, a benchmark of length-matched evidence and
control deletions from three question-answering datasets. Quantitative and qualitative
experiments with five reader models on it show that SELD consistently outperforms
strong and recent baselines across compressors, compression rates and models,
identifying the evidence-removing edit in 83\% of matched pairs with
unchanged answers, against 70\% for the strongest baseline. These results show that evidence loss in compressed contexts can be
detected even when the answer is unchanged.
\end{abstract}

\section{Introduction}
Language models rarely read the full original text anymore. Retrieved
passages are pruned, prompts are shortened by learned token classifiers, and
agents pass condensed notes and memories to each other
\citep{jiang-etal-2023-llmlingua,pan2024llmlingua2,xu2024recomp,chirkova2025provence}.
Each step can drop the evidence that a question depends on, so a deployed
system needs to know, without gold answers, whether a compressed context
still supports the answer. The common safeguard is to ask the model again:
if the answer with the compressed context matches the answer with the full
context, the compression is assumed to be harmless
\citep{manakul-etal-2023-selfcheckgpt,cohenwang2024contextcite}.

\begin{figure*}[t]
\centering
\includegraphics[width=\textwidth]{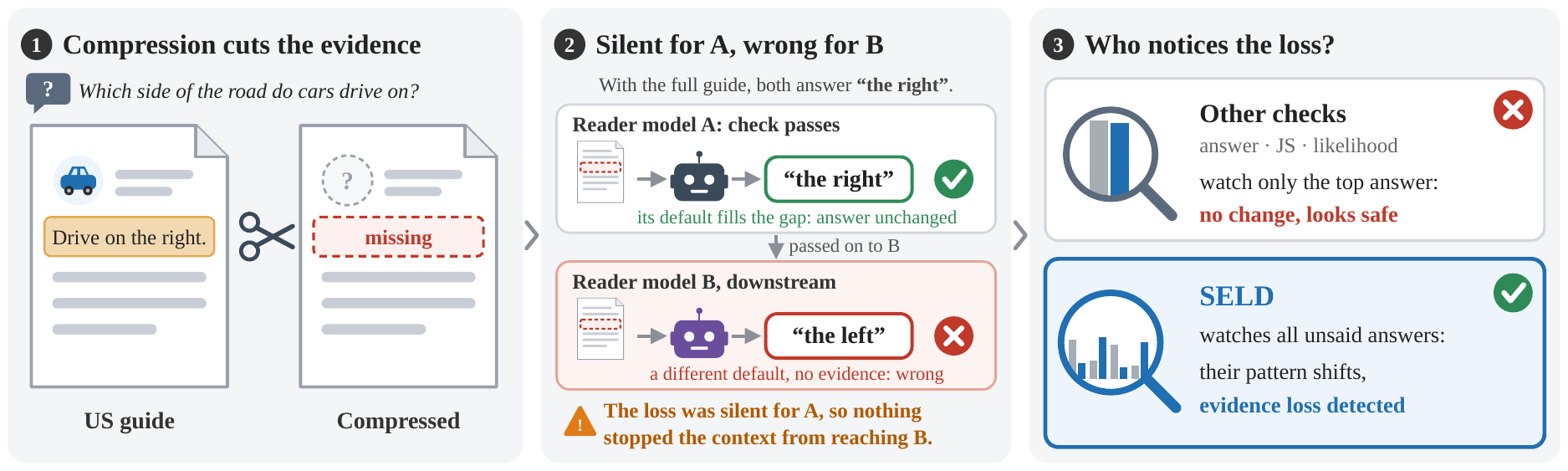}
\caption{\textbf{Silent evidence loss and how SELD detects it}
(illustrative; bars: gray for the original context, blue for the compressed
one). \textbf{(1)} Compression removes the only sentence that states the
answer. \textbf{(2)} With the full guide, both reader models answer \emph{the right}.
After compression, reader model A still answers \emph{the right} from its default,
so an answer check passes; reader model B, reading the same compressed context
downstream, falls back on a different default and answers \emph{the left}.
The loss was silent for A, so nothing stopped the context from reaching B. \textbf{(3)} Checks that look at
the top answer (answer agreement, JS divergence, answer-likelihood drop) see
almost no change; in our tests they change less than when an equally long,
harmless sentence is deleted. SELD compares the reader model's preferences among
all the answers it leaves unsaid, which shift when the evidence is gone. A
real case with measured scores is given in \appref{app:case}.}
\label{fig:overview}
\end{figure*}

This safeguard has a blind spot (\cref{fig:overview}). Suppose a US travel guide
says \emph{Drive on the right} and the question is \emph{Which side of the
road do cars drive on?} After compression removes this sentence, reader model A
still answers \emph{the right} from its default, so the answer check passes.
The compressed context is then passed to reader model B, which also answers
\emph{the right} with the full guide, but now falls back on a different
default and answers \emph{the left}. The answer of A survived; its evidence
did not. We call this
\emph{silent evidence loss} (SEL). It is common: deleting a key sentence and,
separately, an equally long non-key sentence leaves the answer unchanged
both times in 63--78\% of pairs in our development data, and on MuSiQue
\citep{trivedi-etal-2022-musique} about one in five correct answers
reappears with an officially unanswerable context. Such downstream effects are measurable: a key-sentence
deletion that leaves the answer of one model unchanged changes the answers
of \emph{other} models more often than a matched non-key deletion in 6 of 10
cross-family pairs in our data.

That answers can survive missing evidence is known from multi-hop question answering (QA) and
retrieval-augmented generation
\citep{trivedi2020dire,lee2021evidentiality,joren2025sufficient}, and recent
work argues that final answers hide the effect of evidence
\citep{jain2026cuer}. What is missing is a label-free way to \emph{see} SEL
when a context is compressed. Checks based on the answer, such as answer
agreement or the likelihood of the original answer
\citep{cohenwang2024contextcite}, are blind to SEL by construction: they
observe the reader model only through the answer that did not change. Scores that compare output distributions, such as Jensen--Shannon (JS)
divergence \citep{lin1991divergence} or the recent LUMINA \citep{yeh2026lumina}, weight tokens by
probability, so they are dominated by the few most likely tokens and inherit
much of the same blindness. Evaluation is a further obstacle: existing benchmarks do not separate
evidence loss from mere text loss, so a detector can look good simply because removing
more text changes more.

An unchanged answer only means that the same candidate still wins, much
like an election in which the winner stays while the vote counts shift. We
therefore read the reader model's belief over all candidates and ask whether its
pattern, not its winner, has changed. Our contributions are as follows.
\begin{itemize}[leftmargin=*,itemsep=1pt,topsep=2pt,parsep=0pt]
\item \textbf{Problem and benchmark.} We identify and define \emph{silent
evidence loss} (\cref{def:sel}), a frequent failure of context compression
in which key evidence is removed but the answer stays the same. To measure
it, we build \textbf{SEL-Pairs}, a benchmark from three public multi-hop QA
datasets (HotpotQA, 2WikiMultihopQA and MuSiQue) in which every evidence
deletion is matched by an exactly equally long control deletion from the
same context, both for single sentences and inside real outputs of two
compressors, so that a detector cannot win simply because more text was
removed. On it, answer checks miss most silent evidence loss
(\cref{sec:blind}).
\item \textbf{Method.} We show that output divergences mix a change in the
reader model's confidence with a change in the \emph{pattern} of its preferences
(\cref{prop:unsaid}), and find that lost evidence shows in the pattern. SELD
(Silent Evidence Loss Detector) isolates this pattern; it needs no training,
labels or generated answer, and one forward pass (\cref{sec:method}).
\item \textbf{Experiments.} On SEL-Pairs with five reader models, SELD beats
JS divergence by 12.4 points and the answer-likelihood drop by 28.8 points
in a test fixed before data collection (\cref{sec:main}), and beats JS
divergence and an adapted LUMINA detector in every fixed test on real
compressor outputs that could be estimated (\cref{sec:swap}); simpler
explanations do not account for the gains (\cref{sec:notjust}).
\item \textbf{Further findings.} Lost evidence leaves its trace in the
answers the reader model would almost never say: the signal survives removing the
1,000 most likely tokens and reappears in the reader model's hidden state
(\cref{sec:why}). SELD and answer agreement are complementary, and combined
they flag lost and silently lost evidence better than either alone
(\cref{sec:scope}).
\end{itemize}

\section{Related Work}
\paragraph{Compressing contexts and checking the result.}
Compressors range from token pruning \citep{li-etal-2023-compressing,
jiang-etal-2023-llmlingua,jiang-etal-2024-longllmlingua,pan2024llmlingua2}
and question-aware sentence pruning \citep{chirkova2025provence} to learned
and task-aware methods \citep{xu2024recomp,chuang-etal-2024-learning,
liskavets2025cpc,xu-etal-2025-procut,tang-etal-2026-gmsa} and key--value
(KV) cache eviction \citep{li2024snapkv,kim2025kvzip}; they are evaluated
mainly by answer accuracy, and studies ask what information they keep
\citep{lajewska-etal-2025-understanding}. Text-side frameworks verify that
units of meaning survive \citep{trukhina2026commitments}, QAGS, FEQA and
QuestEval check summaries with generated questions
\citep{wang-etal-2020-asking,durmus-etal-2020-feqa,scialom-etal-2021-questeval},
and trained checkers read representations \citep{zeng-etal-2025-towards} or
teach a model to abstain when KV compression evicts evidence
\citep{khodabandehlou2026abstention}. In multi-step agents, concurrent work
finds that compaction silently drops constraints and state, detected by task
outcomes \citep{chen2026governance}, LLM judges
\citep{wang2026lostincompaction,chen2026slipstream} or paired environment
replays \citep{min2026trace}. SELD needs no labels, training or replay and
could serve as their feature.

\paragraph{Answers without evidence.}
Multi-hop QA models answer through disconnected reasoning
\citep{trivedi2020dire}, reader models can be right without evidence
\citep{lee2021evidentiality}, and language models are often correct with
insufficient context \citep{joren2025sufficient}. CUE-R removes and
duplicates single evidence items and argues that the final answer hides
their effect \citep{jain2026cuer}, without controlling prompt length or
comparing output distributions.

\paragraph{Comparing output distributions.}
Context attribution scores the likelihood of a fixed answer under ablated
contexts \citep{cohenwang2024contextcite}. PECoRe and Du et al.\ compare
predictions with and without context by Kullback--Leibler (KL) divergence or mutual information
\citep{sarti2024pecore,du2024context}; LUMINA measures context use through
an embedding-based distance over the most likely tokens
\citep{yeh2026lumina}; distribution-based perturbation tests compare
sampled answers \citep{rauba2025dbpa}; and for quantized models, divergences
of the full next-token distribution reveal changes that accuracy hides
\citep{qamar2026accuracy}. The cosine of first-token logits has served as an ad hoc acceptance
threshold when shortening code documentation \citep{yang2026shortendoc};
we instead derive the score from invariance requirements, relate it to the
divergence family (\cref{prop:unsaid}) and evaluate it as a detector.
Unlike these works, we target the case in which the
answer does not move, test it with length-matched pairs, and show that much
of the useful signal lies outside the tokens that probability-weighted
scores emphasize.

\section{Method}
\label{sec:method}
\begin{figure*}[t]
\centering
\includegraphics[width=\textwidth]{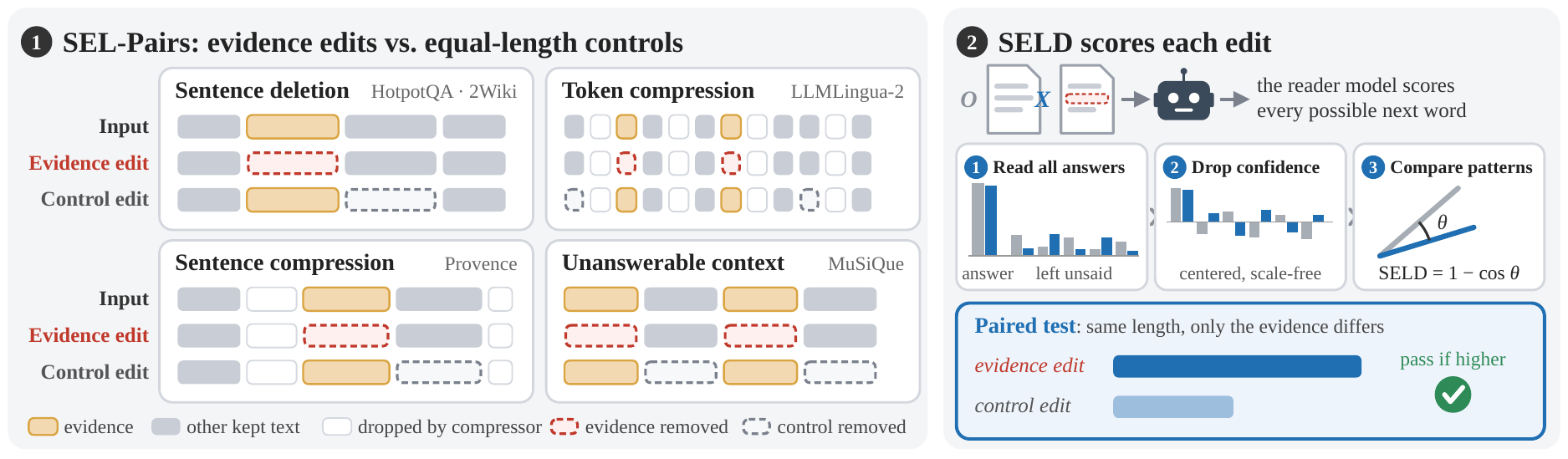}
\caption{\textbf{SEL-Pairs and SELD} (schematic). \textbf{(1)} For each
input, an \emph{evidence edit} (red) removes evidence and a \emph{control
edit} (gray, dashed) removes other text of exactly the same token length:
single-sentence deletion from the original context on HotpotQA and
2WikiMultihopQA (denoted $S$ and $C$ in the text); swaps inside LLMLingua-2
outputs at 25, 50 and 75\% and inside Provence outputs ($B$ and $C$, with
the output as input $A$; LLMLingua-2 swaps are refilled to equal length);
and MuSiQue's official unanswerable context against a control that keeps the
key paragraphs ($U$ and $T$). \textbf{(2)} SELD (step 1) reads the reader model's scores
for every possible next word under $O$ and $X$, so that the answers left
unsaid count; (2) centers them and ignores their scale, so that a mere
change in confidence does not count; and (3) measures the angle between the
two patterns. A check passes a pair when it scores the evidence edit above
its equal-length control.}
\label{fig:method}
\end{figure*}

SELD has two parts: a score that compares the output of the reader model
under the original and the edited context, and a paired protocol that tests whether the
score tracks evidence rather than the amount of text removed. We first
define the phenomenon, then the score, and finally the protocol;
\cref{fig:method} gives an overview.

\paragraph{Silent evidence loss.}
A frozen reader model $M$ answers a question $q$ from a context in its usual
QA prompt; $\hat a_M(q,\cdot)$ is its complete answer
after standard normalization. The original context $O$ contains evidence
$E\subseteq O$ that supports the answer, and is replaced by an edited or
compressed $X$.
\begin{definition}[Silent evidence loss]\label{def:sel}
$X$ exhibits silent evidence loss for $(M,q)$ if (i) evidence is lost,
$E\not\subseteq X$, and (ii) the answer is unchanged,
$\hat a_M(q,X)=\hat a_M(q,O)$, with both answers non-empty and ended by a
native stop.
\end{definition}
The definition depends on evidence annotations and answers only, not on any
score. It does not require the answer to be correct, and it does not require
$X$ to be unanswerable: annotated evidence marks support, not necessity. We
want a score $s(X)$, computed without gold answers or evidence labels, that
is higher when $X$ has lost evidence of $O$.

\paragraph{The SELD score.}
Before generating an answer, the reader model assigns a score to every
token of its vocabulary; we treat this vector as its belief over all
candidate answers, of which the generated answer is only the top entry.
Removing evidence can leave the top entry in place while reshaping the
rest, so a detector of silent evidence loss should (R1) read the whole
belief, (R2) ignore a constant shift of all scores (the log-normalizer) and
a uniform change of sharpness (the overall confidence), and (R3) compare
edits of the same context on one scale. Let $\boldsymbol{\ell}_X\in\mathbb{R}^V$ denote the log-probabilities
that $M$ assigns to the $V$ vocabulary tokens at the first answer position
given $q$ and $X$, and let
$\tilde{\boldsymbol{\ell}}_X=\boldsymbol{\ell}_X-\bar\ell_X\mathbf{1}$, where
$\bar\ell_X$ is the mean of its entries. Centering removes the shift and
normalization removes the scale, which gives
\begin{equation}
\mathrm{SELD}(X)=1-\frac{\langle\tilde{\boldsymbol{\ell}}_X,\tilde{\boldsymbol{\ell}}_O\rangle}{\|\tilde{\boldsymbol{\ell}}_X\|_2\,\|\tilde{\boldsymbol{\ell}}_O\|_2}.
\label{equ:unsaid}
\end{equation}
SELD needs one forward pass per context and no generated answer, labels or
training.
\begin{proposition}\label{prop:unsaid}
(i) For any $c_1,c_2\in\mathbb{R}$ and $\lambda_1,\lambda_2>0$, replacing
$\boldsymbol{\ell}_X$ by $\lambda_1\boldsymbol{\ell}_X+c_1\mathbf{1}$ and
$\boldsymbol{\ell}_O$ by $\lambda_2\boldsymbol{\ell}_O+c_2\mathbf{1}$ leaves
SELD unchanged. Hence SELD does not depend on the softmax temperature and
equals $1-\rho(\mathbf{z}_X,\mathbf{z}_O)$, where $\mathbf{z}$ denotes the
logits and $\rho$ the Pearson correlation. (ii) Let
$\mathbf{a}=\tilde{\boldsymbol{\ell}}_O$, $\mathbf{b}=\tilde{\boldsymbol{\ell}}_X$,
and let $\mathrm{JS}_T$ be the JS divergence between
$\mathrm{softmax}(\mathbf{a}/T)$ and $\mathrm{softmax}(\mathbf{b}/T)$. Then
$\lim_{T\to\infty}8VT^2\,\mathrm{JS}_T=(\|\mathbf{a}\|_2-\|\mathbf{b}\|_2)^2+2\|\mathbf{a}\|_2\|\mathbf{b}\|_2\,\mathrm{SELD}(X)$.
\end{proposition}
By (ii), divergences between output distributions mix a
\emph{confidence} term (overall sharpness) with a \emph{pattern} term, and
at ordinary temperatures they weigh mostly the top few tokens; SELD
isolates the pattern term (proofs in \appref{app:temperature}). Centering is
necessary: log-probabilities share a large negative offset, so the cosine of
uncentered vectors is dominated by this offset and, unlike SELD, varies with
the temperature.

\paragraph{Paired, length-matched evaluation.}
For each context we build two edits that leave the same number of
reader-model tokens: $S$ removes evidence and $C$ removes the same amount of other text.
A check passes a pair if it scores $S$ above $C$, with ties counted as one
half. The \emph{pass rate}, averaged over questions, is our metric; 50\% is
chance, and a score that only tracks how much text was removed is at chance
by construction. The answer is \emph{unchanged} for a pair when
condition (ii) of \cref{def:sel} holds for both $S$ and $C$; this group is
defined from the answers alone, and answer agreement is at chance in it by
definition.

\section{Experimental Setup}
\label{sec:data}
The test sets below form \textbf{SEL-Pairs}. We release them as source
identifiers and edit spans with scripts that rebuild every input.
\paragraph{Test 1: sentence deletion.}
We drew 2,414 questions from the HotpotQA (1,742) and 2WikiMultihopQA (672)
training sets \citep{yang-etal-2018-hotpotqa,ho-etal-2020-constructing};
none shares a paragraph with another selected question or with our
development data. $S$ deletes an annotated supporting sentence and $C$ a
non-supporting sentence of the same paragraph with exactly the same token
length for that reader model. 1,116, 985, 1,186 and 1,116 questions qualify for
the four reader models (4,459 pairs). Before running any model, we fixed the
inputs, code and two main tests: SELD versus JS divergence and versus the
answer-likelihood drop, on pairs with unchanged answers, averaged over
reader models; eight per-reader tests count only if both main tests pass.

\paragraph{Test 2: evidence swaps inside real compressed outputs.}
Sentence deletion is a clean but artificial edit, so we also edit real
compressor outputs. We start from the output $A$ of LLMLingua-2
\citep{pan2024llmlingua2} on the same 2,414 questions. LLMLingua-2 does not
see the question, so it keeps some annotated support words and drops
others. If $A$ kept $k$ support words, $B$ removes exactly these $k$ words,
while $C$ removes $k$ kept non-support words whose compressor scores match
them and keeps every support word of $A$. Both arms refill $k$ words that
the compressor had dropped, chosen by its own scores and mostly shared, so
that $A$, $B$ and $C$ have exactly the same prompt length for each reader model
(\cref{fig:method}). We run this at 50\% and, in a second fixed family, at
25\% and 75\% of the original length. For Provence
\citep{chirkova2025provence}, which keeps or drops whole sentences, $B$
removes one kept support sentence and $C$ one kept non-support sentence, so
that the prompts of $B$ and $C$ have exactly the same length. Granite-3.3-8B-Instruct
\citep{granite2025} serves as a fifth reader model from a fourth family at 50\%.
Construction reads no reader model output; it fails when no exact length match is
found or word alignment is incomplete, never because of answers
(\cref*{app:swap,app:newsettings}). For each setting, the main tests
(SELD versus chance, JS divergence and LUMINA-ECU on $B$ against $C$,
averaged over reader models) and, where planned, per-reader tests were fixed
before any reader model saw $B$ or $C$. The family for 25\% and 75\% also holds
the three tests of an original Provence design that proved not estimable.

\paragraph{Additional datasets.} MuSiQue-Full \citep{trivedi-etal-2022-musique}
pairs each answerable context $O$ with an officially unanswerable context
$U$; a control $T$ keeps every key paragraph and matches the length of $U$
exactly (about 1,550 questions per reader model). Natural outputs of LLMLingua-2
and Provence on 4,831 questions and LongBench \citep{bai-etal-2024-longbench}
with four compressors map the limits of SELD. Earlier HotpotQA and 2Wiki
cohorts shaped the hypothesis (\appref{app:dev}).

\paragraph{Baselines.}
Besides JS divergence of the same vectors, the \emph{answer-likelihood drop}
is the mean per-token log-probability of the original answer of the reader model,
given $O$ minus given $X$ \citep{cohenwang2024contextcite}; \emph{answer
agreement} compares the generated answers. \emph{LUMINA-ECU} adapts the
external context utilization (ECU) score of LUMINA \citep{yeh2026lumina}: at
the first answer position it compares the top-100 tokens under $O$ and $X$,
keeping their probability mass, by a squared maximum mean discrepancy
(MMD$^2$) with a shifted-cosine kernel over the input embeddings of the
reader model (\appref{app:lumina}); it is an adaptation, not a reproduction. Six
\emph{text-only scores} see only the question and the removed text: word
overlap, Jaccard, query recall, BM25 \citep{robertson2009bm25}, BGE cosine
\citep{xiao2024cpack} and the reranker bge-reranker-v2-m3
\citep{chen-etal-2024-m3}.

\paragraph{Metrics, statistics and implementation.}
All tests use the pass rate, MuSiQue with $U$ in place of $S$ and $T$ in
place of $C$, and alarms across questions use the area under the ROC curve
(AUROC). We resample questions within each dataset (100,000 draws) and
report Bonferroni-corrected percentile intervals in brackets; family sizes
are given in captions and in \appref{app:protocol}. There was no external registration; all other analyses
are post hoc and reuse saved scores, and a second implementation reproduced
the reported statistics. The reader models are
Qwen2.5-1.5B-Instruct, Mistral-7B-Instruct-v0.3, Olmo-3-7B-Instruct and
Qwen2.5-14B-Instruct \citep{qwen2025qwen25,jiang2023mistral,olmo3-instruct-card},
named below by family and size (e.g., Qwen2.5-1.5B). They run frozen in BF16 with native
chat templates and no truncation, answer greedily with a 32-token cap, and
we save full-vocabulary log-probabilities in FP32 at the first answer
position. Code, SEL-Pairs and the numerical supplement will be released; all appendices are in the supplementary material.

\section{Results}
\subsection{Prevalence of Silent Evidence Loss}
\label{sec:blind}
On the development data, deleting a key sentence and, separately, its
control leaves the answer unchanged both times for 63--78\% of pairs whose
answers could be read. In Test 1, 66--72\% of qualifying questions have at
least one such pair. On MuSiQue, among
questions that a reader model answers correctly with $O$ and usably with $U$, the
same correct answer reappears with the unanswerable $U$ for 18.3--22.4\% of
questions. Unchanged answers are not a sign of safety either: where the
answers to $O$, $U$ and $T$ all agree, only 15--29\% are correct. When the
answer of one reader model is unchanged, the same key-sentence deletion changes
the answer of another model family more often than the control in 6 of 10
reader-model pairs (post hoc), although a drop in F1 for
that model is established in only 1 of 10. An unchanged answer says little
about the evidence.

\subsection{Detecting Evidence Loss behind Unchanged Answers}
\label{sec:main}
\begin{figure*}[t]
\centering
\includegraphics[width=\textwidth]{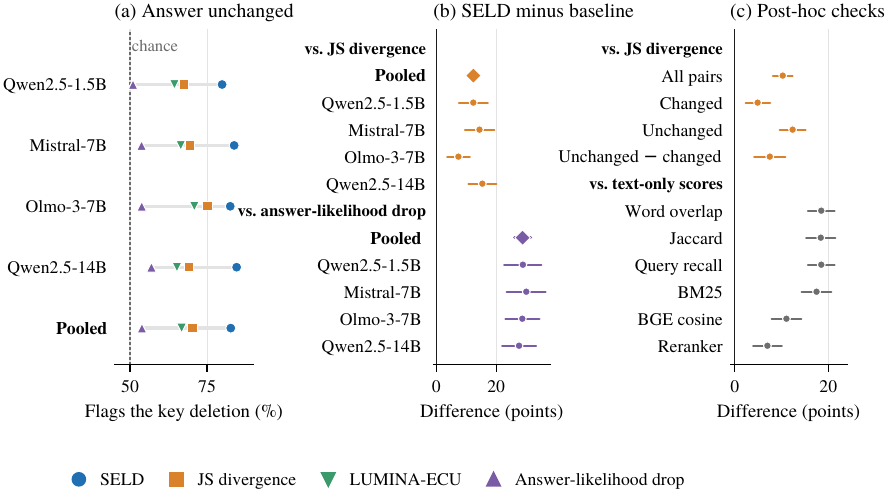}
\caption{\textbf{Test 1: single-sentence deletion on 2,414 new questions.}
(a) How often each check scores the key-sentence deletion above its
equal-length control, among pairs whose answer did not change (50\% is
chance; LUMINA-ECU is post hoc). (b) The two main tests (pooled over
reader models) and eight per-reader tests. (c) Post-hoc checks: SELD minus JS
divergence by answer group and SELD minus six text-only scores among
unchanged answers. Intervals are Bonferroni-corrected within each family
(\appref{app:protocol}).}
\label{fig:main}
\end{figure*}
Both main tests of Test 1 pass (\cref{fig:main}b). Among pairs with
unchanged answers, SELD flags the key-sentence deletion in 82.7\% of
cases, JS divergence in 70.3\% and the answer-likelihood drop in 53.9\%,
close to chance (\cref{fig:main}a). SELD beats JS divergence by
\textbf{12.4 points} [10.4, 14.3] and the answer-likelihood drop by
\textbf{28.8 points} [26.2, 31.5], and all eight per-reader tests pass: the
gain over JS divergence ranges from 7.4 (Olmo-3-7B) to 15.4 points (Qwen2.5-14B).
The answer-likelihood drop scores the answer that did not change, and JS
divergence weighs the same few tokens. LUMINA-ECU is weaker still, 66.7\% (post hoc); SELD
beats it in all eight dataset--reader cells, by 8.6 to 28.2 points among
unchanged answers, with every interval above zero.
LUMINA-ECU compares only the top 100 tokens, so it cannot observe a change
in the pattern over the rest of the vocabulary. A real case in which SELD ranks the key deletion first and the two baselines do not is given in \appref{app:case}.

\subsection{Generalization to Real Compressors}
\label{sec:swap}
\begin{figure*}[t]
\centering
\includegraphics[width=\textwidth]{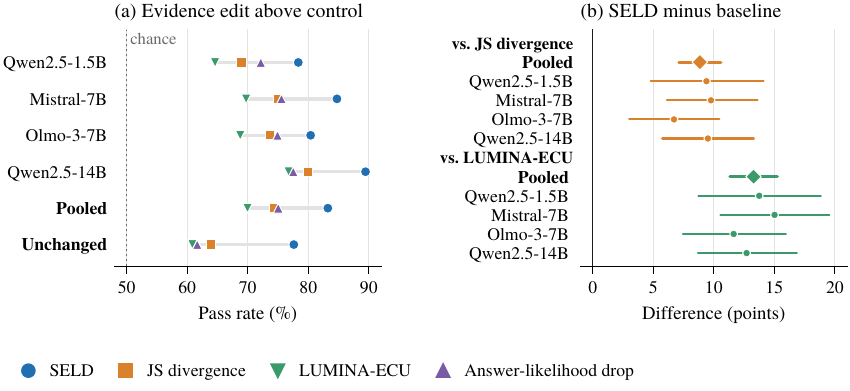}
\caption{\textbf{Test 2 at 50\%: evidence swaps inside real LLMLingua-2
outputs} (construction in \cref{fig:method}). (a) How often each score
ranks the evidence edit above its control, per reader model, pooled, and pooled over questions whose
answers under $O$, $A$, $B$ and $C$ are identical (descriptive). (b) Two of
the three main tests (pooled) and the corresponding eight per-reader tests,
with Bonferroni-corrected intervals; SELD versus chance, +33.2 [31.4, 35.0],
is not shown.}
\label{fig:swap}
\end{figure*}
\begin{table*}[t]
\centering\small
\setlength{\tabcolsep}{5pt}
\begin{tabular}{lrrrrrr}
\toprule
Setting & Pairs & SELD & JS & ECU & SELD $-$ JS & SELD $-$ ECU\\
\midrule
LLMLingua-2 25\% & 6,491 & \textbf{75.9} & 66.3 & 61.9 & +9.6 [7.8, 11.4] & +14.0 [12.0, 16.0]\\
LLMLingua-2 50\% & 3,702 & \textbf{83.2} & 74.4 & 70.0 & +8.8 [7.1, 10.6] & +13.3 [11.3, 15.2]\\
LLMLingua-2 75\% & 801 & \textbf{89.7} & 75.9 & 73.9 & +13.9 [9.5, 18.3] & +15.8 [11.1, 20.6]\\
Provence, sentences & 258 & \textbf{73.3} & 60.7 & 60.9 & +12.7 [4.2, 21.0] & +12.4 [2.2, 22.5]\\
\midrule
Granite-3.3-8B, LLMLingua-2 50\% & 682 & \textbf{83.1} & 71.4 & 62.3 & +11.7 [7.2, 16.3] & +20.8 [15.9, 25.8]\\
\bottomrule
\end{tabular}
\caption{Test 2 across compressors, rates and model families: how often each
score ranks the support-removing edit $B$ above the matched control $C$ (\%,
mean over the four reader models; Granite-3.3-8B is a single fifth reader model) and
differences in points. Pairs counts question--reader pairs; each reader model has
its own exactly token-matched swaps. Intervals are corrected within each fixed family: the
50\%, Provence and Granite-3.3-8B settings for 3 tests each, 25\% and 75\% jointly
for 9. Every per-reader test of the LLMLingua-2 settings passes.}
\label{tab:settings}
\end{table*}
LLMLingua-2 drops words throughout the text rather than whole sentences. At 50\%, SELD
ranks the evidence edit above its equal-length control in \textbf{83.2\%} of
cases, against 74.4\% for JS divergence, 70.0\% for LUMINA-ECU and 75.0\%
for the answer-likelihood drop (\cref{fig:swap}; the last is descriptive). It beats JS divergence by
\textbf{8.8 points} [7.1, 10.6] and LUMINA-ECU by \textbf{13.3 points}
[11.3, 15.2], and all twelve per-reader tests pass. The removed support
words matter: relative to $C$, $B$ lowers the F1 of the reader model by 25--34
points on average (descriptive). Yet for the questions whose answer stays
the same under $O$, $A$, $B$ and $C$, SELD still ranks $B$ above $C$ in
77.6\% of cases, against 63.9\% for JS divergence and 61.7\% for the
answer-likelihood drop, and its lead over JS divergence is 12--15 points for
every reader model, against 5--8 points when the answer changes (descriptive).

The result holds across compressors, rates and model families
(\cref{tab:settings}). Every estimable fixed test passes: at 25\% and 75\%
for all four reader models, for whole-sentence deletions inside Provence outputs,
and for Granite-3.3-8B, a fifth reader model from a fourth family. Exact length matching is a strict requirement of
the evaluation, not of SELD; it is met for 67--69\% of the questions at
25\% and for fewer at higher rates (\appref{app:newsettings}).

\subsection{Ruling Out Alternative Explanations}
\label{sec:notjust}
\paragraph{Text similarity and self-judgment.}
In Test 1, all six text-only scores fall short of SELD among unchanged
answers (\cref{fig:main}c): by 17.5--18.5 points for word-based scores and
BM25, 11.1 for BGE cosine and 7.0 [3.9, 10.1] for the trained reranker. In
Test 2, on the questions where the control removal overlaps the question at
least as much as the support removal, SELD still beats JS divergence by
6.9--13.4 points for every reader model. On the development data, SELD also
beats a prompt-based sufficiency check in which the reader model judges
whether the passages alone answer the question
\citep[cf.][]{joren2025sufficient}, by 7.9--18.1 points in six of eight
dataset--reader cells and ties in the other two (post hoc;
\appref{app:dev}).

\paragraph{Case selection and parametric memory.}
Over all pairs of Test 1, SELD beats JS divergence by 10.3 points [8.1,
12.4]; when the answer changes, the gain shrinks to 4.9 [2.3, 7.6], and it is
7.5 points [4.1, 10.8] larger when the answer is unchanged (for three
reader models, not for Olmo-3-7B). For 751 pairs with unchanged answers whose
deletion removes the only mention of the answer or a bridge entity, the
reader model answers correctly without any context in only 87, so closed-book
recall explains few of them (post hoc). In Test 2 at 50\%, qualifying questions
have fewer support words on HotpotQA and shorter contexts on 2Wiki than the
others, but the lead of SELD over JS divergence holds in every
support-size band with at least 30 questions, by 6.0--16.0 points
(descriptive, \appref{app:selection}).

\paragraph{Beyond annotated supporting sentences.}
On MuSiQue, which labels whether a context can answer at all, SELD scores
the unanswerable $U$ above the length-matched $T$ more often than JS
divergence and the answer-likelihood drop for every reader model, by 7.8 [6.2,
9.5] and 8.7 [6.7, 10.7] points (post hoc, \appref{app:musique}). For
correct answers surviving in $U$, the gain over JS divergence (+9.1
[$-1.5$, 19.6], 141 questions) is not established.

\subsection{Analysis: Pattern versus Confidence}
\label{sec:why}
\begin{figure}[t]
\centering
\includegraphics[width=\columnwidth]{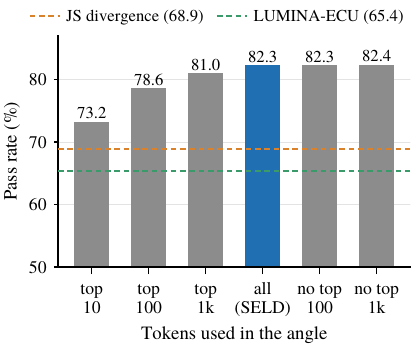}
\caption{\textbf{The signal does not need the top tokens.} Test 1, unchanged
answers: pass rate of the angle computed on subsets of the vocabulary,
ranked by probability under $O$ and re-centered (unweighted mean over the
eight dataset--reader cells; post hoc, descriptive).}
\label{fig:vocab}
\end{figure}
Two ingredients separate SELD from JS divergence: it compares by angle
rather than by probability-weighted divergence, and it uses the whole
vocabulary. \Cref{fig:vocab} isolates both. Computed on the top 10 tokens
only, the angle already passes in 73.2\% of unchanged pairs, above JS
divergence (68.9\%) and LUMINA-ECU (65.4\%) on the same cells, so the way of
comparing matters. More of the vocabulary helps further: 78.6\% with the top
100, 81.0\% with the top 1,000 and 82.3\% with all tokens. Without the top
100 or 1,000 tokens it is unchanged (82.3\% and 82.4\%). What the reader model leaves unsaid carries the trace of the lost evidence.
Raising the softmax temperature of JS divergence from 1 to 16 spreads its
weight and improves its silent-loss pass rate in all eight development combinations,
by 4.1 to 20.6 points, approaching SELD as \cref{prop:unsaid} predicts,
but makes it worse at detecting a \emph{change} of answer in four of them
(\appref{app:temperature}); SELD is the temperature-free limit. The same principle carries over to the reader model's internal state: the
angle between final normalized hidden states, of which the logits are a
linear function, reaches 82.5\% against 82.7\% for SELD in Test 1
(difference +0.2 [$-0.4$, 0.9], post hoc). The signal thus
reflects the reader model's state, not an artifact of the output layer, and
SELD recovers it from the output distribution alone.

\subsection{Complementarity with Answer Checks}
\label{sec:scope}
Compression is performed by the system that holds the original context, so
the reference that SELD needs is available exactly when a check is due,
as in reference-based evaluation of generation. SELD and
answer agreement answer two different questions. To flag outputs
whose answer is lost, answer agreement is the natural check, and for silent
loss it is blind (\cref{tab:twochecks}): the more a check weighs the top
answer, the better it sees a lost answer and the worse it sees silent loss.
\begin{table}[t]
\centering\small
\setlength{\tabcolsep}{3pt}
\begin{tabular}{lrr}
\toprule
Check & Silent loss (Test 1) & Lost answer\\
\midrule
Answer agreement & 50.0 & \textbf{96.7--97.8}\\
Answer-likelihood drop & 53.9 & 81.4--83.9\\
JS divergence & 70.3 & 81.2--83.2\\
SELD & \textbf{82.7} & 71.8--78.6\\
\bottomrule
\end{tabular}
\caption{Two questions, two checks. Left: Test 1 pass rate among unchanged
answers (\%). Right: AUROC for flagging LongBench compressor outputs whose
answer is lost (F1 drops to 0 for a question answered with F1$\geq$50 using
$O$), range over two reader models.}
\label{tab:twochecks}
\end{table}
The two are therefore best used together: averaging the percentile ranks of
SELD and answer agreement, calibrated on one half of the development
questions, beats SELD alone by 2.8 AUROC points [2.2, 3.5] and answer
agreement alone by 16.2 [14.0, 18.6] on the other half, for a target
positive when the answer changed or stayed despite a removed key sentence. SELD adds one forward pass
per candidate (29--30\,ms with Qwen2.5-1.5B; \appref{app:cost}). Ranking contexts of \emph{different} questions
would also need a per-question calibration (\appref{app:natural}).

\section{Conclusion}
In this work, we study silent evidence loss in context compression, where
key evidence is removed while the answer stays the same. We introduce SELD,
a training-free detector that compares a reader model's centered
log-probabilities before and after compression, and SEL-Pairs, a
benchmark of length-matched evidence and control edits. Across five reader
models, compressors and compression rates, SELD identifies the
evidence-removing edit more often than answer-, likelihood- and
divergence-based checks (83\% against 70\% on unchanged answers). SELD
needs the original context and the output distribution; future work will
extend it to multi-step agents, whose compressed histories are reused across
many decisions.

\section*{Limitations}
\textbf{Evidence labels.} We use annotated supporting sentences as evidence.
They mark evidence rather than strict necessity, so a flagged context is not
proven insufficient, and the source of a surviving answer is analysed only
in part (\appref{app:posthoc}).
\textbf{Coverage.} Exact length matching inside real compressor outputs, a
requirement of our evaluation rather than of SELD, covers a subset of
question--reader slots, largest at moderate compression rates; within it the
advantage holds across support sizes
(\cref*{app:newsettings,app:selection}).
\textbf{Scope.} We study English multi-hop QA with open reader models of
1.5B to 14B parameters. SELD reads the reader model's output distribution,
available for open-weight and self-hosted models, and compares each
compressed context with its own original, as a paired diagnostic that
complements answer checks (\cref{sec:scope}).

\section*{Ethical Considerations}
We use public research benchmarks and frozen, openly released models for
offline diagnosis. No human participants were recruited and no new personal
data were collected. Benchmark texts can contain names of public figures; we
did not perform a content-level privacy audit. Asset licenses are listed in
\cref*{app:assets}; the numerical supplement contains
scores, generated answers and opaque identifiers rather than benchmark
source texts or model weights. A check that agrees with the original
context does not certify factual completeness or safety and should not
replace task-specific review in high-stakes settings.

\bibliography{main}

@inproceedings{bai-etal-2024-longbench,
    title = "{L}ong{B}ench: A Bilingual, Multitask Benchmark for Long Context Understanding",
    author = "Bai, Yushi  and
      Lv, Xin  and
      Zhang, Jiajie  and
      Lyu, Hongchang  and
      Tang, Jiankai  and
      Huang, Zhidian  and
      Du, Zhengxiao  and
      Liu, Xiao  and
      Zeng, Aohan  and
      Hou, Lei  and
      Dong, Yuxiao  and
      Tang, Jie  and
      Li, Juanzi",
    editor = "Ku, Lun-Wei  and
      Martins, Andre  and
      Srikumar, Vivek",
    booktitle = "Proceedings of the 62nd Annual Meeting of the Association for Computational Linguistics (Volume 1: Long Papers)",
    month = aug,
    year = "2024",
    address = "Bangkok, Thailand",
    publisher = "Association for Computational Linguistics",
    url = "https://aclanthology.org/2024.acl-long.172/",
    doi = "10.18653/v1/2024.acl-long.172",
    pages = "3119--3137"
}

@inproceedings{lajewska-etal-2025-understanding,
    title = "Understanding and Improving Information Preservation in Prompt Compression for {LLM}s",
    author = "{\L}ajewska, Weronika  and
      Hardalov, Momchil  and
      Aina, Laura  and
      Anna John, Neha  and
      Su, Hang  and
      Marquez, Lluis",
    editor = "Christodoulopoulos, Christos  and
      Chakraborty, Tanmoy  and
      Rose, Carolyn  and
      Peng, Violet",
    booktitle = "Findings of the Association for Computational Linguistics: EMNLP 2025",
    month = nov,
    year = "2025",
    address = "Suzhou, China",
    publisher = "Association for Computational Linguistics",
    url = "https://aclanthology.org/2025.findings-emnlp.949/",
    doi = "10.18653/v1/2025.findings-emnlp.949",
    pages = "17520--17541",
    ISBN = "979-8-89176-335-7"
}

@inproceedings{tang-etal-2026-gmsa,
    title = "{GMSA}: Enhancing Context Compression via Group Merging and Layer Semantic Alignment",
    author = "Tang, Jiwei  and
      Zhang, Zhicheng  and
      Wu, Shunlong  and
      Ye, Jingheng  and
      Bai, Lichen  and
      Wang, Zitai  and
      Lu, Tingwei  and
      Hai, Lin  and
      Zhao, Yiming  and
      Zheng, Hai-Tao  and
      Kim, Hong-Gee",
    editor = "Liakata, Maria  and
      Moreira, Viviane P.  and
      Zhang, Jiajun  and
      Jurgens, David",
    booktitle = "Proceedings of the 64th Annual Meeting of the {A}ssociation for {C}omputational {L}inguistics (Volume 1: Long Papers)",
    month = jul,
    year = "2026",
    address = "San Diego, California, United States",
    publisher = "Association for Computational Linguistics",
    url = "https://aclanthology.org/2026.acl-long.1324/",
    doi = "10.18653/v1/2026.acl-long.1324",
    pages = "28690--28704",
    ISBN = "979-8-89176-390-6"
}

@inproceedings{pan2024llmlingua2,
    title = "{LLML}ingua-2: Data Distillation for Efficient and Faithful Task-Agnostic Prompt Compression",
    author = {Pan, Zhuoshi  and
      Wu, Qianhui  and
      Jiang, Huiqiang  and
      Xia, Menglin  and
      Luo, Xufang  and
      Zhang, Jue  and
      Lin, Qingwei  and
      R{\"u}hle, Victor  and
      Yang, Yuqing  and
      Lin, Chin-Yew  and
      Zhao, H. Vicky  and
      Qiu, Lili  and
      Zhang, Dongmei},
    editor = "Ku, Lun-Wei  and
      Martins, Andre  and
      Srikumar, Vivek",
    booktitle = "Findings of the Association for Computational Linguistics: ACL 2024",
    month = aug,
    year = "2024",
    address = "Bangkok, Thailand",
    publisher = "Association for Computational Linguistics",
    url = "https://aclanthology.org/2024.findings-acl.57/",
    doi = "10.18653/v1/2024.findings-acl.57",
    pages = "963--981"
}

@article{yang2026shortendoc,
    title = {Less is More: {DocString} Compression in Code Generation},
    author = {Yang, Guang and Zhou, Yu and Cheng, Wei and Zhang, Xiangyu and Chen, Xiang and Zhuo, Terry Yue and Zhou, Xin and Liu, Ke and Lo, David and Chen, Taolue},
    journal = {ACM Transactions on Software Engineering and Methodology},
    volume = {35},
    number = {2},
    pages = {1--31},
    month = feb,
    year = {2026},
    doi = {10.1145/3735636},
    url = {https://arxiv.org/abs/2410.22793v3},
    note = {Method description inspected in author manuscript arXiv:2410.22793v3 (June 2025); first preprint October 2024}
}

@inproceedings{liskavets2025cpc,
    title = {Prompt Compression with Context-Aware Sentence Encoding for Fast and Improved {LLM} Inference},
    author = {Liskavets, Barys and Ushakov, Maxim and Roy, Shuvendu and Klibanov, Mark and Etemad, Ali and Luke, Shane K.},
    booktitle = {Proceedings of the AAAI Conference on Artificial Intelligence},
    volume = {39},
    pages = {24595--24604},
    year = {2025},
    doi = {10.1609/aaai.v39i23.34639},
    url = {https://ojs.aaai.org/index.php/AAAI/article/view/34639}
}

@inproceedings{chuang-etal-2024-learning,
    title = "Learning to Compress Prompt in Natural Language Formats",
    author = "Chuang, Yu-Neng  and
      Xing, Tianwei  and
      Chang, Chia-Yuan  and
      Liu, Zirui  and
      Chen, Xun  and
      Hu, Xia",
    editor = "Duh, Kevin  and
      Gomez, Helena  and
      Bethard, Steven",
    booktitle = "Proceedings of the 2024 Conference of the North American Chapter of the Association for Computational Linguistics: Human Language Technologies (Volume 1: Long Papers)",
    month = jun,
    year = "2024",
    address = "Mexico City, Mexico",
    publisher = "Association for Computational Linguistics",
    url = "https://aclanthology.org/2024.naacl-long.429/",
    doi = "10.18653/v1/2024.naacl-long.429",
    pages = "7756--7767"
}

@inproceedings{xu-etal-2025-procut,
    title = "{P}ro{C}ut: {LLM} Prompt Compression via Attribution Estimation",
    author = "Xu, Zhentao  and
      Li, Fengyi  and
      Chen, Albert C.  and
      Wang, Xiaofeng",
    editor = "Potdar, Saloni  and
      Rojas-Barahona, Lina  and
      Montella, Sebastien",
    booktitle = "Proceedings of the 2025 Conference on Empirical Methods in Natural Language Processing: Industry Track",
    month = nov,
    year = "2025",
    address = "Suzhou (China)",
    publisher = "Association for Computational Linguistics",
    url = "https://aclanthology.org/2025.emnlp-industry.20/",
    doi = "10.18653/v1/2025.emnlp-industry.20",
    pages = "285--309",
    ISBN = "979-8-89176-333-3"
}

@inproceedings{yang-etal-2018-hotpotqa,
  title     = {{HotpotQA}: A Dataset for Diverse, Explainable Multi-hop Question Answering},
  author    = {Yang, Zhilin and Qi, Peng and Zhang, Saizheng and Bengio, Yoshua and Cohen, William and Salakhutdinov, Ruslan and Manning, Christopher D.},
  booktitle = {Proceedings of the 2018 Conference on Empirical Methods in Natural Language Processing},
  year      = {2018}
}

@inproceedings{wang-etal-2020-asking,
  title={Asking and Answering Questions to Evaluate the Factual Consistency of Summaries},
  author={Wang, Alex and Cho, Kyunghyun and Lewis, Mike},
  booktitle={Proceedings of the 58th Annual Meeting of the Association for Computational Linguistics},
  year={2020}, pages={5008--5020}, doi={10.18653/v1/2020.acl-main.450},
  url={https://aclanthology.org/2020.acl-main.450/}
}

@inproceedings{scialom-etal-2021-questeval,
  title={{QuestEval}: Summarization Asks for Fact-based Evaluation},
  author={Scialom, Thomas and Dray, Paul-Alexis and Lamprier, Sylvain and Piwowarski, Benjamin and Staiano, Jacopo and Wang, Alex and Gallinari, Patrick},
  booktitle={Proceedings of the 2021 Conference on Empirical Methods in Natural Language Processing},
  year={2021}, url={https://aclanthology.org/2021.emnlp-main.529/}
}

@misc{olmo3-instruct-card,
  title={{Olmo 3 7B Instruct}: Model Card},
  author={{Allen Institute for AI}}, year={2025},
  url={https://huggingface.co/allenai/Olmo-3-7B-Instruct},
  note={Evaluated revision 6e5971d}
}

@article{qwen2025qwen25,
  title   = {{Qwen2.5} Technical Report},
  author  = {{Qwen Team}},
  journal = {arXiv preprint arXiv:2412.15115},
  year    = {2024}
}

@article{jiang2023mistral,
  title   = {{Mistral 7B}},
  author  = {Jiang, Albert Q. and Sablayrolles, Alexandre and Mensch, Arthur and Bamford, Chris and Chaplot, Devendra Singh and de las Casas, Diego and Bressand, Florian and Lengyel, Gianna and Lample, Guillaume and Saulnier, Lucile and others},
  journal = {arXiv preprint arXiv:2310.06825},
  year    = {2023}
}

@inproceedings{jiang-etal-2023-llmlingua,
  title     = {{LLML}ingua: Compressing Prompts for Accelerated Inference of Large Language Models},
  author    = {Jiang, Huiqiang and Wu, Qianhui and Lin, Chin-Yew and Yang, Yuqing and Qiu, Lili},
  booktitle = {Proceedings of the 2023 Conference on Empirical Methods in Natural Language Processing},
  year      = {2023},
  doi       = {10.18653/v1/2023.emnlp-main.825}
}

@inproceedings{jiang-etal-2024-longllmlingua,
  title     = {{LongLLMLingua}: Accelerating and Enhancing {LLM}s in Long Context Scenarios via Prompt Compression},
  author    = {Jiang, Huiqiang and Wu, Qianhui and Luo, Xufang and Li, Dongsheng and Lin, Chin-Yew and Yang, Yuqing and Qiu, Lili},
  booktitle = {Proceedings of the 62nd Annual Meeting of the Association for Computational Linguistics (Volume 1: Long Papers)},
  year      = {2024}
}

@inproceedings{li-etal-2023-compressing,
  title     = {Compressing Context to Enhance Inference Efficiency of Large Language Models},
  author    = {Li, Yucheng and Dong, Bo and Guerin, Frank and Lin, Chenghua},
  booktitle = {Proceedings of the 2023 Conference on Empirical Methods in Natural Language Processing},
  year      = {2023}
}

@inproceedings{xu2024recomp,
  title     = {{RECOMP}: Improving Retrieval-Augmented {LM}s with Context Compression and Selective Augmentation},
  author    = {Xu, Fangyuan and Shi, Weijia and Choi, Eunsol},
  booktitle = {International Conference on Learning Representations},
  year      = {2024}
}

@inproceedings{ho-etal-2020-constructing,
  title     = {Constructing A Multi-hop {QA} Dataset for Comprehensive Evaluation of Reasoning Steps},
  author    = {Ho, Xanh and Duong Nguyen, Anh-Khoa and Sugawara, Saku and Aizawa, Akiko},
  booktitle = {Proceedings of the 28th International Conference on Computational Linguistics},
  pages     = {6609--6625},
  year      = {2020},
  doi       = {10.18653/v1/2020.coling-main.580}
}

@inproceedings{durmus-etal-2020-feqa,
  title     = {{FEQA}: A Question Answering Evaluation Framework for Faithfulness Assessment in Abstractive Summarization},
  author    = {Durmus, Esin and He, He and Diab, Mona},
  booktitle = {Proceedings of the 58th Annual Meeting of the Association for Computational Linguistics},
  pages     = {5055--5070},
  year      = {2020},
  doi       = {10.18653/v1/2020.acl-main.454}
}

@article{hinton2015distilling,
  title = {Distilling the Knowledge in a Neural Network},
  author = {Hinton, Geoffrey and Vinyals, Oriol and Dean, Jeff},
  journal = {arXiv preprint arXiv:1503.02531},
  year = {2015}
}

@inproceedings{zeng-etal-2025-towards,
  title = {Towards Knowledge Checking in Retrieval-augmented Generation: A Representation Perspective},
  author = {Zeng, Shenglai and Zhang, Jiankun and Li, Bingheng and Lin, Yuping and Zheng, Tianqi and Everaert, Dante and Lu, Hanqing and Liu, Hui and Liu, Hui and Xing, Yue and Cheng, Monica Xiao and Tang, Jiliang},
  booktitle = {Proceedings of the 2025 Conference of the Nations of the Americas Chapter of the Association for Computational Linguistics: Human Language Technologies (Volume 1: Long Papers)},
  year = {2025},
  month = apr,
  address = {Albuquerque, New Mexico},
  publisher = {Association for Computational Linguistics},
  pages = {2952--2969},
  doi = {10.18653/v1/2025.naacl-long.151},
  url = {https://aclanthology.org/2025.naacl-long.151/}
}

@article{trivedi-etal-2022-musique,
  title = {{M}u{S}i{Q}ue: Multihop Questions via Single-hop Question Composition},
  author = {Trivedi, Harsh and Balasubramanian, Niranjan and Khot, Tushar and Sabharwal, Ashish},
  journal = {Transactions of the Association for Computational Linguistics},
  volume = {10},
  pages = {539--554},
  year = {2022}
}

@inproceedings{cohenwang2024contextcite,
  title = {{ContextCite}: Attributing Model Generation to Context},
  author = {Cohen-Wang, Benjamin and Shah, Harshay and Georgiev, Kristian and Madry, Aleksander},
  booktitle = {Advances in Neural Information Processing Systems},
  volume = {37},
  year = {2024}
}

@inproceedings{joren2025sufficient,
  title = {Sufficient Context: A New Lens on Retrieval Augmented Generation Systems},
  author = {Joren, Hailey and Zhang, Jianyi and Ferng, Chun-Sung and Juan, Da-Cheng and Taly, Ankur and Rashtchian, Cyrus},
  booktitle = {International Conference on Learning Representations},
  year = {2025}
}

@inproceedings{manakul-etal-2023-selfcheckgpt,
  title = {{S}elf{C}heck{GPT}: Zero-Resource Black-Box Hallucination Detection for Generative Large Language Models},
  author = {Manakul, Potsawee and Liusie, Adian and Gales, Mark},
  booktitle = {Proceedings of the 2023 Conference on Empirical Methods in Natural Language Processing},
  year = {2023}
}

@inproceedings{xiao2024cpack,
  title = {{C-Pack}: Packed Resources for General {C}hinese Embeddings},
  author = {Xiao, Shitao and Liu, Zheng and Zhang, Peitian and Muennighoff, Niklas and Lian, Defu and Nie, Jian-Yun},
  booktitle = {Proceedings of the 47th International ACM SIGIR Conference on Research and Development in Information Retrieval},
  year = {2024}
}

@inproceedings{chen-etal-2024-m3,
  title = {{M3-Embedding}: Multi-Linguality, Multi-Functionality, Multi-Granularity Text Embeddings Through Self-Knowledge Distillation},
  author = {Chen, Jianlyu and Xiao, Shitao and Zhang, Peitian and Luo, Kun and Lian, Defu and Liu, Zheng},
  booktitle = {Findings of the Association for Computational Linguistics: ACL 2024},
  year = {2024}
}

@article{robertson2009bm25,
  title = {The Probabilistic Relevance Framework: {BM25} and Beyond},
  author = {Robertson, Stephen and Zaragoza, Hugo},
  journal = {Foundations and Trends in Information Retrieval},
  volume = {3},
  number = {4},
  pages = {333--389},
  year = {2009}
}

@article{lin1991divergence,
  title = {Divergence Measures Based on the {S}hannon Entropy},
  author = {Lin, Jianhua},
  journal = {IEEE Transactions on Information Theory},
  volume = {37},
  number = {1},
  pages = {145--151},
  year = {1991}
}

@article{khodabandehlou2026abstention,
  title = {Compression-Aware Abstention: Teaching {LLMs} to Refuse When {KV}-Compression Masks Remove Answer Evidence},
  author = {Khodabandehlou, Mohammadali and Krishnamachari, Bhaskar},
  journal = {arXiv preprint arXiv:2608.29934},
  year = {2026}
}

@article{trukhina2026commitments,
  title = {Compress the Context, Keep the Commitments: A Formal Framework for Verifiable {LLM} Context Compression},
  author = {Trukhina, Natalia and Vashkelis, Vadim},
  journal = {arXiv preprint arXiv:2605.17304},
  year = {2026}
}

@inproceedings{yeh2026lumina,
  title = {{LUMINA}: Detecting Hallucinations in {RAG} System with Context--Knowledge Signals},
  author = {Yeh, Min-Hsuan and Li, Yixuan and Mallick, Tanwi},
  booktitle = {International Conference on Learning Representations},
  year = {2026},
  note = {arXiv:2509.21875}
}

@inproceedings{chirkova2025provence,
  title = {Provence: Efficient and Robust Context Pruning for Retrieval-Augmented Generation},
  author = {Chirkova, Nadezhda and Formal, Thibault and Nikoulina, Vassilina and Clinchant, St{\'e}phane},
  booktitle = {International Conference on Learning Representations},
  year = {2025}
}

@inproceedings{li2024snapkv,
  title = {{SnapKV}: {LLM} Knows What You are Looking for Before Generation},
  author = {Li, Yuhong and Huang, Yingbing and Yang, Bowen and Venkitesh, Bharat and Locatelli, Acyr and Ye, Hanchen and Cai, Tianle and Lewis, Patrick and Chen, Deming},
  booktitle = {Advances in Neural Information Processing Systems},
  volume = {37},
  year = {2024}
}

@inproceedings{kim2025kvzip,
  title = {{KVzip}: Query-Agnostic {KV} Cache Compression with Context Reconstruction},
  author = {Kim, Jang-Hyun and Kim, Jinuk and Kwon, Sangwoo and Lee, Jae W. and Yun, Sangdoo and Song, Hyun Oh},
  booktitle = {Advances in Neural Information Processing Systems},
  volume = {38},
  year = {2025}
}

@inproceedings{trivedi2020dire,
  title = {Is Multihop {QA} in {DiRe} Condition? Measuring and Reducing Disconnected Reasoning},
  author = {Trivedi, Harsh and Balasubramanian, Niranjan and Khot, Tushar and Sabharwal, Ashish},
  booktitle = {Proceedings of the 2020 Conference on Empirical Methods in Natural Language Processing},
  pages = {8846--8863},
  year = {2020}
}

@inproceedings{lee2021evidentiality,
  title = {Robustifying Multi-hop {QA} through Pseudo-Evidentiality Training},
  author = {Lee, Kyungjae and Hwang, Seung-won and Han, Sang-eun and Lee, Dohyeon},
  booktitle = {Proceedings of the 59th Annual Meeting of the Association for Computational Linguistics and the 11th International Joint Conference on Natural Language Processing},
  pages = {6110--6119},
  year = {2021}
}

@article{jain2026cuer,
  title = {{CUE-R}: Beyond the Final Answer in Retrieval-Augmented Generation},
  author = {Jain, Siddharth and Vedam, Venkat Narayan},
  journal = {arXiv preprint arXiv:2604.05467},
  year = {2026}
}

@inproceedings{sarti2024pecore,
  title = {Quantifying the Plausibility of Context Reliance in Neural Machine Translation},
  author = {Sarti, Gabriele and Chrupa{\l}a, Grzegorz and Nissim, Malvina and Bisazza, Arianna},
  booktitle = {International Conference on Learning Representations},
  year = {2024}
}

@inproceedings{du2024context,
  title = {Context versus Prior Knowledge in Language Models},
  author = {Du, Kevin and Sn{\ae}bjarnarson, V{\'e}steinn and Stoehr, Niklas and White, Jennifer C. and Schein, Aaron and Cotterell, Ryan},
  booktitle = {Proceedings of the 62nd Annual Meeting of the Association for Computational Linguistics},
  year = {2024}
}

@inproceedings{rauba2025dbpa,
  title = {Statistical Hypothesis Testing for Auditing Robustness in Language Models},
  author = {Rauba, Paulius and Wei, Qiyao and van der Schaar, Mihaela},
  booktitle = {Proceedings of the 42nd International Conference on Machine Learning},
  year = {2025}
}

@article{qamar2026accuracy,
  title = {Accuracy is Not Enough: A Divergence-Based Approach to Evaluate Fidelity Loss in Quantized {LLMs}},
  author = {Qamar, Shahzeb and Sparrenberg, Lorenz and Bauckhage, Christian and Rababah, Baha and Leung, Carson and Kantarcioglu, Murat and Akcora, Cuneyt Gurcan and Sifa, Rafet},
  journal = {arXiv preprint arXiv:2609.07664},
  year = {2026}
}

@misc{granite2025,
  title = {Granite 3.3 8{B} Instruct: Model Card},
  author = {{IBM Research}},
  year = {2025},
  url = {https://huggingface.co/ibm-granite/granite-3.3-8b-instruct}
}

@article{chen2026slipstream,
  title={{Slipstream}: Trajectory-Grounded Compaction Validation for Long-Horizon Agents},
  author={Chen, Zhuofu and Pan, Rui and Dai, Yinwei and Netravali, Ravi},
  journal={arXiv preprint arXiv:2605.08580},
  year={2026}
}

@article{wang2026lostincompaction,
  title={Lost in Compaction: Evaluating Side-Constraint Loss under Context Compaction},
  author={Wang, Zhiqi and Zhang, Yichi and Lee, Dongwon and Yang, Yuchen},
  journal={arXiv preprint arXiv:2608.11242},
  year={2026}
}

@article{chen2026governance,
  title={Governance Decay: How Context Compaction Silently Erases Safety Constraints in Long-Horizon {LLM} Agents},
  author={Chen, Shiyang},
  journal={arXiv preprint arXiv:2606.22528},
  year={2026}
}

@article{min2026trace,
  title={Toward Reliable Context Compression for Long-Horizon Agents: An Empirical Study of Execution Instability},
  author={Min, Guanghui and Wu, Liang and Darbari, Mayank and Chen, Chen and Hong, Liangjie},
  journal={arXiv preprint arXiv:2608.06503},
  year={2026}
}

\clearpage
\renewcommand{\CL}{\mathrm{SELD}}
\newcommand{\partheading}[1]{\par\vspace{10pt}\noindent{\large\bfseries #1}\par\vspace{4pt}}
\renewcommand{\topfraction}{0.9}\renewcommand{\bottomfraction}{0.8}
\renewcommand{\textfraction}{0.07}\renewcommand{\floatpagefraction}{0.85}
\renewcommand{\dbltopfraction}{0.9}\renewcommand{\dblfloatpagefraction}{0.85}
\setcounter{topnumber}{3}\setcounter{bottomnumber}{2}\setcounter{totalnumber}{5}\setcounter{dbltopnumber}{3}
\setcounter{table}{0}\setcounter{figure}{0}\setcounter{equation}{0}\setcounter{footnote}{0}
\renewcommand{\theHtable}{supp.\arabic{table}}
\renewcommand{\theHfigure}{supp.\arabic{figure}}
\renewcommand{\theHequation}{supp.\arabic{equation}}
\twocolumn[{\centering\Large\bfseries Supplementary Material\par\vspace{1.5em}}]

\appendix
\renewcommand{\thesection}{\AlphAlph{\value{section}}}
\renewcommand{\thetable}{S\arabic{table}}
\renewcommand{\thefigure}{S\arabic{figure}}
\section*{Guide to the Supplement}
This supplement documents the evidence behind each claim of the main paper.
\Cref{tab:guide} maps every claim to the appendices that support it and
states the status of each analysis: \emph{fixed} tests were set before the
data they evaluate, \emph{post hoc} analyses were added afterwards with
their own multiple-comparison corrections, and \emph{descriptive} results
carry no inferential claim. Part~I covers the main results, Part~II the
mechanism, Part~III the scope and practical use, and Part~IV development
history and reproducibility. Tables and figures are numbered S1, S2, \dots;
references to the main paper are marked as such.

\paragraph{Notation.} A \emph{reader} is the reader model of the main paper. Qwen2.5-1.5B, Mistral-7B, Olmo-3-7B and Qwen2.5-14B denote the
instruction-tuned checkpoints Qwen2.5-1.5B-Instruct, Mistral-7B-Instruct-v0.3,
Olmo-3-7B-Instruct and Qwen2.5-14B-Instruct (revisions in \cref{app:assets});
the external judge in \cref{app:natural} is Qwen2.5-32B-Instruct. SELD is the score of the main paper (one minus the
cosine of centered log-probabilities); JS is Jensen--Shannon divergence,
$\Own$ the answer-likelihood drop, $D$ the cosine distance of final hidden
states and ECU the LUMINA adaptation. \emph{Preserved} or \emph{silent}
pairs are those whose answer did not change. Figures carried over from development runs show SELD under its working
label CL.

\begin{table}[h]
\centering\footnotesize
\setlength{\tabcolsep}{2pt}
\begin{tabular}{>{\raggedright\arraybackslash}p{0.47\columnwidth}>{\raggedright\arraybackslash}p{0.27\columnwidth}>{\raggedright\arraybackslash}p{0.2\columnwidth}}
\toprule
Claim (main paper) & Appendices & Status\\
\midrule\addlinespace[1pt]
SEL is frequent; answer checks are blind to it (Sec.~\ref*{sec:blind}) & \cref*{app:protocol,app:posthoc} & fixed, descriptive\\
SELD beats JS, answer likelihood and LUMINA on Test~1 (Sec.~\ref*{sec:main}) & \cref*{app:protocol,app:lumina,app:posthoc,app:case} & fixed; ECU post hoc\\
The advantage holds at 14B scale & \cref*{app:scale} & post hoc, fixed family\\
It carries over to real compressors (Sec.~\ref*{sec:swap}) & \cref*{app:swap,app:newsettings,app:selection} & fixed; selection descriptive\\
It is not explained by simpler factors (Sec.~\ref*{sec:notjust}) & \cref*{app:posthoc,app:musique} & post hoc\\
Mechanism and Proposition~\ref*{prop:unsaid} (Sec.~\ref*{sec:method}, \ref*{sec:why}) & \cref*{app:temperature,app:vocab,app:hiddenD} & proof; post hoc\\
Complementary to answer checks (Sec.~\ref*{sec:scope}) & \cref*{app:natural,app:real,app:cost} & post hoc, descriptive\\
How the hypothesis was developed & \cref*{app:dev,app:confirmation,app:near-length,app:near-length-qwen} & fixed (dev), post hoc\\
Assets, prompts and reproducibility & \cref*{app:impl,app:assets} & --\\
\bottomrule
\end{tabular}
\caption{Where each claim of the main paper is supported.}
\label{tab:guide}
\end{table}
\FloatBarrier
\partheading{Part I: Main Results in Detail}

\section{Confirmatory Protocol and Coverage}
\label{app:protocol}
\paragraph{Inputs.}
The fixed inventory has 2,414 native questions (HotpotQA train 1,742;
2WikiMultihopQA train 672). Selection used only source text: a question
qualifies for a reader before any hashing, and a greedy pass over a fixed
hash order then removes questions that share a paragraph title or body with
an already selected question or with development data. No model output was
read. Each eligible reader keeps every selected support/control reference.
Controls are non-supporting sentences of the same paragraph whose deletion
removes exactly as many tokens as the support deletion under the tokenizer
of that reader; the fixed \texttt{relaxed\_one} fallback lets the
content-word overlap of the control with the question be at most one
lower than that of the support sentence. Qwen2.5-1.5B and Qwen2.5-14B share a tokenizer and therefore the
same references, but remain separate readers.

\paragraph{Collection.}
Each eligible reader produces, for $O$, every $S$ and $C$, and an empty
context $E$: greedy generation with the native stop tokens and a 32-token
answer cap, the full-vocabulary first-answer-position log-probabilities, the
final hidden state, teacher-forced gold likelihood and the teacher-forced
likelihood of its own $O$ answer. No prompt is truncated.

\paragraph{Answer layers.}
Answers are normalized with the official LongBench rule (lower case,
removal of ASCII punctuation and English articles, whitespace collapse) on
the complete decoded response. Empty responses, responses that hit the cap
and responses without a native stop are \emph{unknown}; they are never
counted as answer-changed.

\paragraph{Statistics.}
100,000 bootstrap draws resample questions within each dataset, shared by
all readers and comparisons. Primary intervals are Bonferroni-corrected for
two tests and reader-level intervals for eight; the reader-level tests count
only if both primary tests pass. A planning simulation with an 8-point
effect gave a power of about 0.93 under its assumptions.

\Cref{tab:coverage} reports the coverage of Test 1 and the absolute pass rates among unchanged answers.
\begin{table}[htbp]
\centering\small
\setlength{\tabcolsep}{2pt}
\begin{tabular}{lrrrrr}
\toprule
Reader & Elig.\ q & Refs & Pres.\ q & SELD & JS\\
\midrule
Qwen2.5-1.5B & 1,116 & 1,126 & 770 & 79.87 & 67.53\\
Mistral-7B & 985 & 997 & 666 & 83.78 & 69.37\\
Olmo-3-7B & 1,186 & 1,210 & 786 & 82.51 & 75.13\\
Qwen2.5-14B & 1,116 & 1,126 & 801 & 84.58 & 69.23\\
Pooled & -- & 4,459 & -- & 82.69 & 70.31\\
\bottomrule
\end{tabular}
\caption{Coverage and absolute ranking accuracy (\%) in the
answer-preserving layer. The own-answer drop reaches 50.97, 53.75, 53.80,
56.96 and 53.87\%. Absolute values are descriptive; the questions of
different readers overlap and must not be summed.}
\label{tab:coverage}
\end{table}

\begin{table}[htbp]
\centering\small
\resizebox{\columnwidth}{!}{%
\begin{tabular}{lrr}
\toprule
& SELD$-$JS & SELD$-\Own$\\
\midrule
Pooled (family 2) & 12.37 [10.43, 14.31] & 28.81 [26.17, 31.47]\\
Qwen2.5-1.5B (family 8) & 12.34 [7.53, 17.20] & 28.90 [22.68, 35.08]\\
Mistral-7B & 14.41 [9.57, 19.38] & 30.03 [23.51, 36.49]\\
Olmo-3-7B & 7.38 [3.61, 11.19] & 28.71 [23.03, 34.44]\\
Qwen2.5-14B & 15.36 [10.68, 20.01] & 27.62 [21.96, 33.28]\\
\bottomrule
\end{tabular}}
\caption{All ten fixed confirmatory comparisons (points, Bonferroni 95\%
intervals).}
\label{tab:confirm-all}
\end{table}

\section{Test 2: Evidence Swaps in Compressed Outputs}
\label{app:swap}
\paragraph{Source.}
$A$ is the unmodified output of the official LLMLingua-2 compressor at a
50\% rate for each of the 2,414 questions of Test 1, given the full context
without the question. The word-level decisions of the compressor were replayed
exactly from the saved selection mask; its continuous word scores were
re-collected with the same frozen classifier, because the original run
saved only the binary decisions.

\paragraph{Construction.}
A word belongs to the support set if all of its characters lie inside one
annotated supporting sentence and to the editable non-support set if all
lie inside one non-supporting sentence; titles and separators are
protected. Let $K$ be the kept and $U$ the dropped words of $A$, and $k$ the
number of kept support words. $B$ removes all $k$ kept support words; $C$
removes $k$ kept non-support words chosen by an ordered minimum-$L_1$
matching of compressor scores to the removed support words, and keeps every
support word of $A$. Both arms then add back words from the dropped
non-support pool, in decreasing compressor score, and a deterministic
bounded local search (beam 4, depth 8, at most 256 complete prompts per arm)
swaps single words until the complete native QA prompts of $A$, $B$ and $C$
have exactly the same number of tokens for that reader. No support word is
ever added back, nothing is padded or truncated, and no reader output,
score, answer or gold label is read during construction. On average the two
arms share 91\% of their added words (Qwen2.5-1.5B), and for Qwen2.5-1.5B $k$ has
a median of 21 words.

\paragraph{Coverage.}
Of the 9,656 question--reader slots, 3,702 qualify (Qwen2.5-1.5B 873, Mistral-7B
839, Olmo-3-7B 1,117, Qwen2.5-14B 873). The other slots are recorded as undefined:
the bounded length search was exhausted (3,638 slot-level cases), the word
alignment to the source text was incomplete (2,096), a kept word crossed a
support boundary (116), the original output could not be replayed (92) or
there were too few non-support words (12). Construction never reads answers,
so qualification does not depend on correctness. Alignment failures fall
mostly on 2Wiki, where only 61--81 questions per reader qualify; no
question has identical $A$, $B$ and $C$ texts for all four readers.

\paragraph{Statistics.}
The three main endpoints are the reader-averaged pass rate of SELD minus
50\%, minus JS divergence and minus LUMINA-ECU on $B$ versus $C$, each
computed on the questions where all three scores are finite. 100,000
bootstrap draws resample questions within each dataset,
shared across readers; tails are $0.05/6$ for the three main tests and
$0.05/24$ for the twelve per-reader tests, which count only if all three
main tests pass.

\Cref{tab:swap} gives the per-reader results of Test 2.
\begin{table}[htbp]
\centering\small
\setlength{\tabcolsep}{2.4pt}
\resizebox{\columnwidth}{!}{%
\begin{tabular}{lrrrrrr}
\toprule
& \multicolumn{4}{c}{All qualifying questions} & \multicolumn{2}{c}{Fixed tests}\\
\cmidrule(lr){2-5}\cmidrule(lr){6-7}
Reader & SELD & JS & ECU & Own & $-$JS & $-$ECU\\
\midrule
Qwen2.5-1.5B & 78.4 & 69.0 & 64.6 & 72.2 & 9.4 [4.8, 14.1] & 13.7 [8.7, 18.8]\\
Mistral-7B & 84.7 & 75.0 & 69.7 & 75.6 & 9.8 [6.1, 13.6] & 15.0 [10.5, 19.6]\\
Olmo-3-7B & 80.4 & 73.7 & 68.8 & 74.9 & 6.7 [3.0, 10.4] & 11.6 [7.4, 16.0]\\
Qwen2.5-14B & 89.5 & 80.0 & 76.7 & 77.5 & 9.5 [5.7, 13.3] & 12.7 [8.7, 16.9]\\
Pooled & 83.2 & 74.4 & 70.0 & 75.0 & 8.8 [7.1, 10.6] & 13.3 [11.3, 15.2]\\
\midrule
\multicolumn{7}{l}{\emph{Answers identical under $O$, $A$, $B$, $C$ (descriptive)}}\\
Qwen2.5-1.5B (250) & 71.6 & 56.4 & 54.4 & 59.6 & 15.2 & 17.2\\
Mistral-7B (240) & 79.6 & 64.2 & 59.2 & 57.1 & 15.4 & 20.4\\
Olmo-3-7B (288) & 76.7 & 64.6 & 61.5 & 66.3 & 12.2 & 15.3\\
Qwen2.5-14B (326) & 82.5 & 70.6 & 68.4 & 63.7 & 12.0 & 14.1\\
\bottomrule
\end{tabular}}
\caption{Test 2: how often each score ranks $B$ above $C$ (\%), and SELD
minus a baseline (points). Fixed tests: corrected for 3 (pooled) and 12
(per reader). Own is descriptive and computed where the original answer of the reader
is defined.}
\label{tab:swap}
\end{table}

\paragraph{Further descriptive results.}
\emph{By dataset.} On HotpotQA SELD beats JS divergence by 6.0--9.8 points
across readers; on the small 2Wiki sets the difference is positive for all
four readers (4.5--24.2 points; 61--81 questions). \emph{Lexical overlap.}
On questions where the control removal overlaps the question at least as much
as the support removal (197--260 per reader), SELD passes in 66.8, 78.2,
76.2 and 80.2\% of cases against 59.9, 68.5, 68.5 and 66.8\% for JS
divergence. \emph{Utility.} Relative to $C$, $B$ lowers F1 against the gold
answer by 25.2, 33.0, 32.0 and 34.0 points on average (with $B$ below $C$
for 34--46\% of questions) and the gold-answer
log-likelihood by 1.46, 2.38, 2.36 and 3.35 nats per content token; these
do not certify that every $B$ is unanswerable. \emph{Hidden state.} The
cosine distance of the final hidden state ranks $B$ above $C$ about as often
as SELD (SELD minus it: $-0.5$ to $+1.5$ points over all qualifying
questions, $-1.7$ to $+2.4$ among unchanged answers).

\section{Test 2 at Further Rates, Provence and Granite}
\label{app:newsettings}
\paragraph{Design.}
The LLMLingua-2 settings at 25\% and 75\% reuse the construction of the 50\%
setting (\cref{app:swap}) on the outputs of the same compressor at the other
rates. Their main tests (SELD versus chance, JS divergence and LUMINA-ECU,
averaged over the four readers) form one family of nine together with a
planned Provence setting that used the Punkt sentence splitter; that setting
had 11--12 qualifying questions per reader, below the fixed minimum of 30,
so its three tests are reported as not estimable and the family was not
re-sized. The 24 per-reader tests of the two estimable settings form a
family of 36 with the twelve not-estimable ones. The separate Provence
setting uses the official sentence boundaries of the benchmark, mapped onto
the tokens that Provence kept: $B$ deletes one kept supporting sentence and
$C$ one kept non-supporting sentence, with exactly equal native prompt
length of $B$ and $C$ per reader ($A$ is not length-matched to them). On
2Wiki every control is in the same paragraph; on HotpotQA most controls
(8--11 of 13--17 per reader) come from fully non-supporting paragraphs.
Qualifying questions come mostly from 2Wiki (41--57 per reader against
13--17 from HotpotQA). Granite-3.3-8B-Instruct is analysed as a single
additional reader at 50\% with its own family of three; the extension plan
had listed Llama-3.1-8B or Gemma-2-9B as candidates, and we used the openly
licensed Granite checkpoint instead. All families use
100,000 question resamples stratified by dataset; the definitions
were fixed before any reader saw $B$ or $C$.

\paragraph{LLMLingua-2 at 25\% and 75\%.}
\Cref{tab:rates} lists the per-reader pass rates. All six main tests pass:
SELD minus JS divergence is +9.6 [7.8, 11.4] at 25\% and +13.9 [9.5, 18.3]
at 75\%, and SELD minus LUMINA-ECU is +14.0 [12.0, 16.0] and +15.8 [11.1,
20.6] (corrected for 9). All 24 per-reader tests pass (corrected for 36);
for example, SELD minus JS divergence is +8.27 [4.12, 12.46] for Qwen2.5-1.5B
and +10.51 [6.92, 14.13] for Mistral-7B at 25\%. At 75\%, only 5 2Wiki
questions per reader qualify for Qwen2.5-1.5B and Mistral-7B. Where the
hidden-state angle $D$ is reported (Qwen2.5-1.5B and Mistral-7B), it is within
1.7 points of SELD (descriptive).

\begin{table}[htbp]
\centering\small
\setlength{\tabcolsep}{2.4pt}
\resizebox{\columnwidth}{!}{%
\begin{tabular}{llrrrrr}
\toprule
Rate & Reader & $N$ (coverage) & SELD & JS & ECU & $D$\\
\midrule
25\% & Qwen2.5-1.5B & 1,608 (66.6\%) & 72.9 & 64.6 & 60.1 & 72.2\\
25\% & Mistral-7B & 1,617 (67.0\%) & 77.1 & 66.5 & 62.5 & 77.2\\
25\% & Olmo-3-7B & 1,658 (68.7\%) & 72.6 & 64.7 & 59.9 & --\\
25\% & Qwen2.5-14B & 1,608 (66.6\%) & 81.2 & 69.3 & 65.0 & --\\
75\% & Qwen2.5-1.5B & 180 (7.5\%) & 91.7 & 76.1 & 71.7 & 90.0\\
75\% & Mistral-7B & 160 (6.6\%) & 94.4 & 76.9 & 75.0 & 94.4\\
75\% & Olmo-3-7B & 281 (11.6\%) & 84.0 & 74.4 & 74.0 & --\\
75\% & Qwen2.5-14B & 180 (7.5\%) & 88.9 & 76.1 & 75.0 & --\\
\bottomrule
\end{tabular}}
\caption{LLMLingua-2 at 25\% and 75\%: qualifying questions out of 2,414
and pass rates (\%). Every SELD comparison passes its corrected test. $D$
is shown where reported in the descriptive readout.}
\label{tab:rates}
\end{table}

\paragraph{Provence, sentence deletion.}
Qualifying questions are 66, 54, 72 and 66 for the four readers (2.2--3.0\%
of the questions). The four-reader means are 73.3\% for SELD, 60.7\% for JS
divergence and 60.9\% for LUMINA-ECU; all three main tests pass (SELD minus
chance +23.3 [15.6, 30.5], minus JS divergence +12.7 [4.2, 21.0], minus
LUMINA-ECU +12.4 [2.2, 22.5], corrected for 3). Per reader (descriptive),
SELD, JS divergence and LUMINA-ECU are tied at 65.9\% for Qwen2.5-1.5B, and
SELD leads for Mistral-7B (76.9 against 63.9 and 62.0), Olmo-3-7B (77.1
against 59.0 and 60.4) and Qwen2.5-14B (73.5 against 53.8 and 55.3). Among
questions with unchanged answers, only 17--39 per reader remain, and for
Qwen2.5-1.5B SELD is below JS divergence (55.9 against 67.6\%); we therefore
make no claim about silent loss for this setting.

\paragraph{Granite.}
682 questions qualify (28.3\%; HotpotQA 618, 2Wiki 64). SELD reaches
83.1\%, JS divergence 71.4\% and LUMINA-ECU 62.3\%; all three tests pass
(minus chance +33.1 [29.7, 36.5], minus JS divergence +11.7 [7.2, 16.3],
minus LUMINA-ECU +20.8 [15.9, 25.8], corrected for 3). On HotpotQA SELD
and JS divergence reach 84.5\% and 71.7\%; on the 64 2Wiki questions they
nearly tie (70.3\% and 68.8\%). Among the 93 questions with unchanged
answers, SELD, JS divergence and LUMINA-ECU reach 78.5\%, 58.1\% and
44.1\% (descriptive). The hidden-state angle reaches 83.3\%.

\section{Test 2 Selection Check}
\label{app:selection}
Only 3,702 of 9,656 question--reader slots qualify at 50\%. Construction
reads no answers, but qualifying and non-qualifying questions differ. On
HotpotQA, qualifying questions have fewer supporting sentences
(standardized mean difference $-0.40$ to $-0.42$) and, where the number of
kept support words $k$ is recorded, fewer support words ($-0.59$ to $-0.68$).
On 2Wiki, qualifying questions have much shorter contexts ($-0.93$ to
$-0.99$). We fixed five support-size bands before the analysis ($k$ = 1--16,
17--32, 33--64, 65--128, 129+). In the first three bands every reader has at
least 30 questions, and SELD is above both JS divergence and LUMINA-ECU in
all twelve reader--band cells, by 6.0--16.0 points over JS divergence. The
two largest bands hold 1--3 questions per reader or none and are kept as
such. The rank correlation between $k$ and the SELD difference between $B$
and $C$ is only 0.06--0.15, so we do not claim a dose--response relation.
These are descriptive analyses; they show that the advantage holds across
support sizes among qualifying questions, not that the selection is
harmless.

\section{LUMINA-ECU Adaptation and Results}
\label{app:lumina}
\paragraph{Definition.}
LUMINA \citep{yeh2026lumina} scores how much a model uses its external
context by comparing the output distribution with and without the context,
measuring a kernel distance between the two over token embeddings. We
adapt its external-context-utilization (ECU) score to the first answer
position of our setting: for $O$ and the edited context $X$ separately we
take the top-100 tokens of the reader, keep their original probabilities (not
renormalized), and compute the squared maximum mean discrepancy between the
two weighted token sets with a shifted-cosine kernel over the input
embedding matrix of the reader. Larger values mean a larger change. The definition
was fixed on development data before it was applied to Test 1 and Test 2. It
follows the official code where possible but is an adaptation rather than a
reproduction of the full method, which also uses information about the
internal knowledge of the model.

\paragraph{Results.}
\Cref{tab:lumina} lists SELD minus LUMINA-ECU in the exam pass rate for
every dataset--reader cell of Test 1 (post hoc). All sixteen differences for
all pairs and for unchanged answers have corrected intervals above zero. On
pairs whose answer changed, the differences are smaller, several intervals
include zero and one point is negative (Qwen2.5-14B on 2Wiki, $-3.6$). The point estimates of JS divergence are also above LUMINA-ECU in every
cell of these two groups, but by only 1.6 to 5.3 points.

\begin{table}[htbp]
\centering\small
\setlength{\tabcolsep}{2.4pt}
\resizebox{\columnwidth}{!}{%
\begin{tabular}{llrrrr}
\toprule
& & \multicolumn{2}{c}{Test 1 (post hoc)} & \multicolumn{2}{c}{Development}\\
\cmidrule(lr){3-4}\cmidrule(lr){5-6}
Data & Reader & All & Unchanged & All & Unchanged\\
\midrule
Hotpot & Qwen2.5-1.5B & 10.8 & 14.4 & 11.7 & 14.4\\
Hotpot & Mistral-7B & 14.0 & 16.9 & 12.1 & 13.4\\
Hotpot & Olmo-3-7B & 7.8 & 8.6 & 11.3 & 14.6\\
Hotpot & Qwen2.5-14B & 12.2 & 14.5 & 15.9 & 20.3\\
2Wiki & Qwen2.5-1.5B & 15.8 & 17.5 & 13.2 & 15.4\\
2Wiki & Mistral-7B & 16.0 & 18.1 & 17.1 & 21.3\\
2Wiki & Olmo-3-7B & 15.7 & 16.7 & 11.9 & 11.3\\
2Wiki & Qwen2.5-14B & 25.2 & 28.2 & 17.3 & 20.6\\
\bottomrule
\end{tabular}}
\caption{SELD minus LUMINA-ECU (points) in the exam pass rate. Every
interval excludes zero (Bonferroni over the 48 comparisons of each source; 100,000
resamples).}
\label{tab:lumina}
\end{table}

\section{Post-hoc Analyses of the Confirmatory Cohort}
\label{app:posthoc}
All analyses below reuse the saved confirmatory scores, resample the same
fixed question inventory within each dataset (100,000 draws each), were defined after the confirmatory result was known, and
form their own families.

\paragraph{Answer layers.}
\Cref{tab:layers} lists the 40-comparison family (pooled and four readers,
two baselines, four layers). The unknown layer is small; none of its ten
intervals has a strictly positive lower limit. The direct difference between
the preserved and changed layers (family of ten) is 7.49 [4.12, 10.81] for
SELD$-$JS pooled and 9.03, 8.96, 1.34 and 10.63 points for the four readers
(Olmo-3-7B crosses zero); for SELD$-\Own$ it is 23.47 [19.49, 27.42] pooled and
positive for all readers. Because conditioning on unchanged answers can
handicap answer-based baselines, we treat SELD$-$JS as the primary
interpretation and SELD$-\Own$ as supporting.

\begin{table}[htbp]
\centering\small
\resizebox{\columnwidth}{!}{%
\begin{tabular}{llrr}
\toprule
Layer & Reader & SELD$-$JS & SELD$-\Own$\\
\midrule
All & Pooled & 10.26 [8.14, 12.36] & 21.67 [18.83, 24.52]\\
 & Qwen2.5-1.5B & 9.68 [5.37, 13.96] & 22.09 [16.69, 27.48]\\
 & Mistral-7B & 11.98 [7.75, 16.35] & 23.00 [17.27, 28.71]\\
 & Olmo-3-7B & 6.87 [3.64, 10.15] & 20.93 [16.06, 25.83]\\
 & Qwen2.5-14B & 12.50 [8.27, 16.79] & 20.65 [15.59, 25.68]\\
\midrule
Changed & Pooled & 4.88 [2.31, 7.58] & 5.35 [2.72, 8.08]\\
 & Qwen2.5-1.5B & 3.30 [$-$2.50, 9.15] & 5.71 [0.00, 11.60]\\
 & Mistral-7B & 5.45 [0.79, 10.96] & 6.91 [1.80, 12.78]\\
 & Olmo-3-7B & 6.04 [1.96, 10.68] & 5.90 [1.45, 10.83]\\
 & Qwen2.5-14B & 4.73 [$-$0.68, 10.50] & 2.87 [$-$1.89, 8.00]\\
\midrule
Preserved & Pooled & 12.37 [9.58, 15.15] & 28.81 [25.07, 32.61]\\
\bottomrule
\end{tabular}}
\caption{SELD advantages by answer layer (points; post-hoc Bonferroni family
of 40). Preserved-layer reader values equal the confirmatory points of
\cref{tab:confirm-all} with wider intervals.}
\label{tab:layers}
\end{table}

\paragraph{Text-relevance baselines.}
All six scores were computed for all 5,104 sentence requests with no
missing values (BGE inputs at most 112 tokens, reranker pairs at most 155;
no truncation). Higher means more relevant to the question. BM25 uses each
own sentences of the question as the corpus ($k_1=1.2$, $b=0.75$). Word count and
query recall induce identical rankings within a question, so the six scores
are not six independent tests.

\Cref{tab:text} lists the six text-only scores against SELD.
\begin{table}[htbp]
\centering\small
\setlength{\tabcolsep}{3pt}
\begin{tabular}{lrrr}
\toprule
Text score & $R$ (\%) & SELD$-$score & Bonf.\ 6 interval\\
\midrule
Word overlap & 64.18 & 18.50 & [15.62, 21.34]\\
Jaccard & 64.29 & 18.40 & [15.26, 21.53]\\
Query recall & 64.18 & 18.50 & [15.62, 21.34]\\
BM25 & 65.20 & 17.49 & [14.28, 20.65]\\
BGE cosine & 71.64 & 11.05 & [7.87, 14.22]\\
Reranker & 75.71 & 6.98 & [3.92, 10.07]\\
\midrule
SELD & 82.69 & -- & --\\
\bottomrule
\end{tabular}
\caption{Answer-preserving layer, pooled over readers. Over all references
the reranker reaches 75.40\% against 85.15\% for SELD. Of 24 reader-level
differences, 23 have positive pointwise intervals; Qwen2.5-1.5B against the
reranker is $+2.40$ [$-1.16$, 5.98].}
\label{tab:text}
\end{table}

\paragraph{Lexical strata.}
Strata are defined by content-word overlap of the deleted sentences with the
question, without gold answers or scores. Counts of reader--dataset cells
(out of 8) with positive pointwise 95\% intervals, answer-preserving layer:
support overlap $\le$ control overlap, 6 (SELD$-$JS) and 8 (SELD$-\Own$); equal
overlap, 6 and 8; zero overlap for both sentences, 4 and 5. In the
answer-changed layer the corresponding counts are much lower (2--3 of 8,
and 0 in the zero-overlap stratum), and small 2Wiki cells for Mistral-7B and
Qwen2.5-14B have negative points. All 424 cells of the layer and lexical
analyses, including 95 without a usable interval, are in the supplement.

\paragraph{Temperature and geometry.}
\begin{table}[tbp]
\centering\footnotesize
\setlength{\tabcolsep}{2pt}
\begin{tabular}{lrrr}
\toprule
Dataset/reader & Exam & Answer-change & Angle$-$size\\
\midrule
Hotpot/Qwen2.5-1.5B & \textbf{+7.6} & \textbf{$-$5.5} & \textbf{+16.3}\\
Hotpot/Mistral-7B & \textbf{+10.9} & $-$3.2 & \textbf{+17.3}\\
Hotpot/Olmo-3-7B & \textbf{+4.1} & \textbf{$-$5.4} & \textbf{+17.1}\\
Hotpot/Qwen2.5-14B & \textbf{+9.3} & $-$3.8 & \textbf{+16.1}\\
2Wiki/Qwen2.5-1.5B & \textbf{+13.2} & \textbf{$-$18.5} & \textbf{+24.9}\\
2Wiki/Mistral-7B & \textbf{+14.6} & $-$10.0 & \textbf{+18.9}\\
2Wiki/Olmo-3-7B & \textbf{+13.2} & \textbf{$-$13.6} & \textbf{+15.7}\\
2Wiki/Qwen2.5-14B & \textbf{+20.6} & +0.0 & \textbf{+25.6}\\
\bottomrule
\end{tabular}
\caption{Spreading the weight of JS divergence (temperature 16 minus 1) on
the development data: change in exam pass rate and in AUROC for detecting
an answer change; last column: angle part minus size part of the score
difference (points; post hoc, corrected for 40 tests; bold excludes zero).}
\label{tab:temp}
\end{table}
\Cref{tab:temp} summarizes the 40 paired contrasts. For KL and JS, raising
temperature from 1 to 16 increases ranking accuracy in all eight panels.
The AUROC for detecting an answer change falls significantly only for Qwen2.5-1.5B
and Olmo-3-7B on both datasets; for Mistral-7B and Qwen2.5-14B the intervals cross
zero, and Qwen2.5-14B on 2Wiki is $+0.02$ points.

\paragraph{Automatic evidence checks.}
Three automatic checks probe what the support deletions remove; none is a
semantic-sufficiency label. (i) A surface rule marks a reference positive
when the deletion removes the only occurrence of the answer string or of a
title of the bridge entity: 1,588 positive, 1,301 negative and 1,570 undecidable
of 4,459 references (no official aliases are available). (ii) Closed-book
answers: among surface-positive, answer-preserving references, only 87 are
answered correctly without context, while 656 are answered incorrectly, so
most preserved answers are not simply recalled. In the small subset that is
surface-positive, closed-book correct and answer-preserving, SELD ranks above
JS for all four readers on HotpotQA (11--39 questions each; descriptive).
(iii) Renaming entities in 299 earlier questions for two Qwen2.5
readers did not establish an overall behavioural effect (difference in
differences $+2.09$ and $+1.05$ points, intervals crossing zero). Finally,
when the answer of one reader is preserved, the same support deletion changes
the answer of another family more often than the control in 6 of 10 directed
reader pairs (family of 20), but lowers the gold F1 of that reader significantly
in only 1 of 10.

\paragraph{Second prompt.}
A reworded QA instruction for Qwen2.5-1.5B on its 311 eligible confirmatory
2Wiki questions (315 references; 945 conditions) gives $R$ of 77.33\% (SELD)
and 63.67\% (JS) over all references, a difference of 13.67 [7.72, 19.61]
(single pointwise interval, 100,000 question draws). Among the 27
answer-changed questions, SELD and JS tie and SELD is 3.70 points below the
own-answer drop.

\section{MuSiQue}
\label{app:musique}
We use the official MuSiQue-Full dev file and pair the answerable ($O$) and
unanswerable ($U$) record of each of the 2,417 ids. With $D=O\setminus U$
($k$ removed paragraphs, supporting or not) and $A=U\setminus O$ (added
paragraphs), the control is $T=(O\setminus C)\cup A$, where $C$ is a set of
$k$ non-supporting paragraphs of $O$. $T$ keeps every annotated supporting
paragraph, has the same added paragraphs and the same number of paragraphs
as $U$, and is searched (at most 128 candidates in a fixed order) for an
exact match of the token length of $U$ under the tokenizer of each reader. Paragraphs
are presented in a fixed hash order shared by $O$, $U$ and $T$. $R$ counts
how often a score ranks $U$ above $T$. All $O\cup U$ questions form a single
component of shared paragraphs, so we report points only. The MuSiQue
authors report that 45 of 50 manually checked Full instances agree with the
official labels; the labels are strong but not error-free.

\Cref{tab:musique-app} gives the MuSiQue differences for every reader and answer group.
\begin{table}[htbp]
\centering\small
\setlength{\tabcolsep}{2.5pt}
\begin{tabular}{lrrrr}
\toprule
Reader & All JS & All Own & Pres.\ JS & Pres.\ Own\\
\midrule
Qwen2.5-1.5B & +7.78 & +8.89 & +7.37 & +11.23\\
Mistral-7B & +6.58 & +7.69 & +6.18 & +18.55\\
Olmo-3-7B & +6.82 & +9.29 & +11.08 & +20.06\\
Qwen2.5-14B & +10.15 & +8.86 & +16.31 & +17.54\\
Macro & +7.83 & +8.68 & +10.23 & +16.84\\
\bottomrule
\end{tabular}
\caption{MuSiQue: SELD minus JS and minus own-answer drop (points,
descriptive). Eligible questions 1,556/1,536/1,570/1,556. Answer-changed
macro values are $+7.01$ and $+6.96$. The hidden-state angle $D$ is close to
SELD (macro $R$ 82.97 vs.\ 83.21\%).}
\label{tab:musique-app}
\end{table}

\Cref{tab:musique-answers} counts the correct answers that survive in the unanswerable context.
\begin{table}[htbp]
\centering\footnotesize
\setlength{\tabcolsep}{3pt}
\resizebox{\columnwidth}{!}{%
\begin{tabular}{lrrr}
\toprule
Reader & $O$ correct, $U$ usable & Same answer in $U$ & \%\\
\midrule
Qwen2.5-1.5B & 252 & 46 & 18.3\\
Mistral-7B & 292 & 58 & 19.9\\
Olmo-3-7B & 286 & 64 & 22.4\\
Qwen2.5-14B & 460 & 103 & 22.4\\
\bottomrule
\end{tabular}}
\caption{Correct answers that survive in the officially unanswerable
context. Exact match of complete normalized answers; answers that hit the
cap are excluded, not counted as changed.}
\label{tab:musique-answers}
\end{table}

\paragraph{Question-level intervals.}
A post-hoc analysis resampled the 2,417 MuSiQue ids with replacement
(100,000 draws; the same counts for all readers), and corrected 20
comparisons together: 8 per-reader and 2 mean comparisons on all qualifying
questions, and the same 10 on the subset of questions answered correctly
under $O$ whose complete normalized answer reappears under $U$. The subset
does not require the answer to $T$ to be complete or unchanged. Questions
share paragraphs, so these intervals assume exchangeable questions and are
approximate.

\Cref{tab:musique-subset} gives the question-level intervals for this subset.
\begin{table}[htbp]
\centering\small
\setlength{\tabcolsep}{2pt}
\resizebox{\columnwidth}{!}{%
\begin{tabular}{lrrrrr}
\toprule
Reader & $n$ & SELD / JS / Own (\%) & SELD$-$JS & SELD$-\Own$\\
\midrule
Qwen2.5-1.5B & 46 & 50.0 / 54.4 / 52.2 & $-$4.3 [$-$28.3, 18.9] & $-$2.2 [$-$31.4, 27.0]\\
Mistral-7B & 58 & 70.7 / 63.8 / 58.6 & 6.9 [$-$18.3, 31.8] & 12.1 [$-$15.2, 38.9]\\
Olmo-3-7B & 64 & 78.1 / 60.9 / 56.3 & 17.2 [$-$1.5, 36.2] & 21.9 [$-$2.3, 45.6]\\
Qwen2.5-14B & 103 & 69.9 / 53.4 / 62.1 & 16.5 [1.0, 32.5] & 7.8 [$-$11.2, 26.8]\\
Mean & -- & 67.2 / 58.1 / 57.3 & 9.1 [$-$1.5, 19.6] & 9.9 [$-$3.7, 23.1]\\
\midrule
All, mean & -- & -- & 7.8 [6.2, 9.5] & 8.7 [6.7, 10.7]\\
\bottomrule
\end{tabular}}
\caption{MuSiQue, correct answers that survive in $U$ (271 reader--question
observations from 141 distinct ids): rate of scoring $U$ above $T$ and
paired differences (points; Bonferroni-20 intervals). Uncorrected 95\%
intervals for the subset means are [2.2, 15.9] and [1.1, 18.5]. The last
row repeats the mean over all qualifying questions; all eight per-reader
comparisons there are positive after correction.}
\label{tab:musique-subset}
\end{table}

\section{14B Scale Extension}
\label{app:scale}
This post-hoc extension uses Qwen2.5-14B on the full Hotpot (740 questions,
768 references) and 2Wiki (1,621 questions, 2,072 references) cohorts;
inference uses the 457 and 563 questions outside the largest source
component. No answer generation is added.

Each dataset uses 100,000 component bootstrap draws, and the ten fixed
comparisons form one Bonferroni family. Both scale criteria pass: $D$ and
SELD each exceed chance and JS. The separate SELD--KL contrasts also pass.

\begin{figure}[tbp]\centering
\includegraphics[width=\columnwidth]{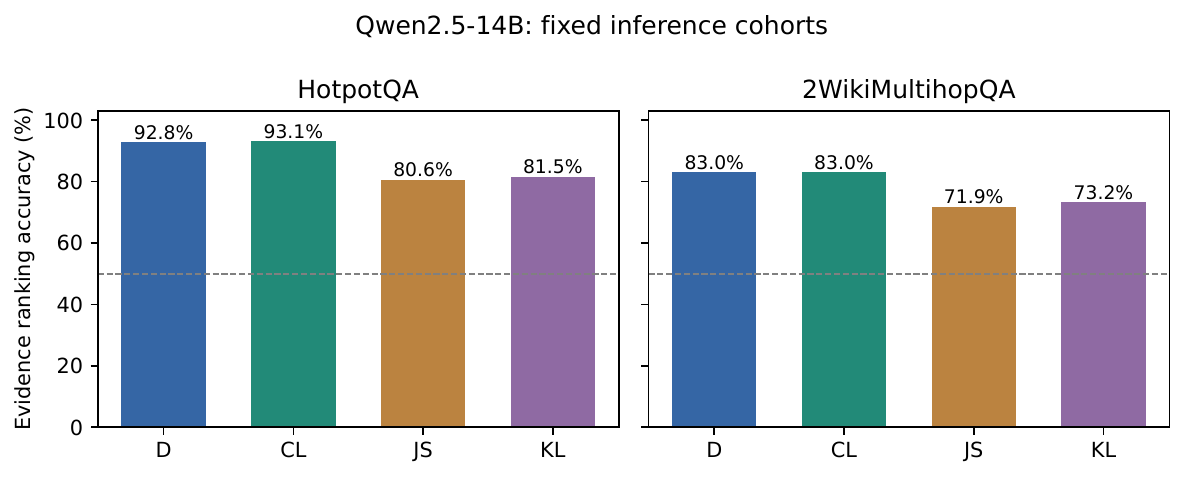}
\caption{14B evidence ordering on the fixed inference cohorts. These
descriptive accuracies are distinct from generated-answer F1.}\label{fig:b2-ranking}
\end{figure}

\begin{figure}[tbp]\centering
\includegraphics[width=\columnwidth]{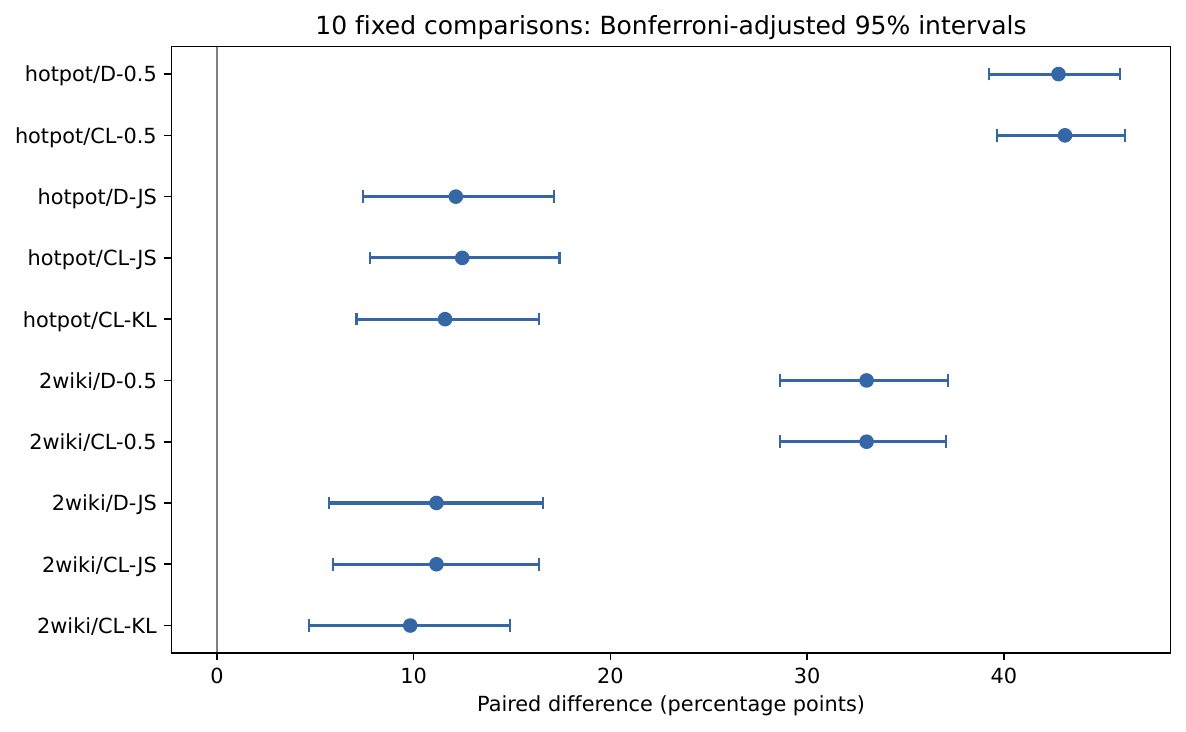}
\caption{All ten fixed scale-extension comparisons with joint adjusted
95\% intervals. No generated-answer utility claim follows from these tests.}\label{fig:b2-comparisons}
\end{figure}

\begin{table}[tbp]\centering\footnotesize
\setlength{\tabcolsep}{3pt}
\caption{Complete scale-extension contrasts.}
\resizebox{\columnwidth}{!}{%
\begin{tabular}{lrrr}
\toprule
Comparison & Difference (points) & Adjusted lower & Adjusted upper\\\midrule
Hotpot/D-0.5 & +42.8 & +39.2 & +45.9\\
Hotpot/SELD-0.5 & +43.1 & +39.6 & +46.2\\
Hotpot/D-JS & +12.1 & +7.4 & +17.1\\
Hotpot/SELD-JS & +12.5 & +7.8 & +17.4\\
Hotpot/SELD-KL & +11.6 & +7.1 & +16.4\\
2Wiki/D-0.5 & +33.0 & +28.6 & +37.2\\
2Wiki/SELD-0.5 & +33.0 & +28.6 & +37.1\\
2Wiki/D-JS & +11.2 & +5.7 & +16.6\\
2Wiki/SELD-JS & +11.2 & +5.9 & +16.4\\
2Wiki/SELD-KL & +9.8 & +4.7 & +14.9\\
\bottomrule\end{tabular}}
\end{table}
\Cref{fig:b2-ranking,fig:b2-comparisons} show the 14B ordering and the ten fixed comparisons.

\section{Case Study}
\label{app:case}
We describe a real Test 1 case for Olmo-3-7B on 2WikiMultihopQA. The
question is \emph{Were both Herbert Lang and Jason Goss, born in the same
place?}; the official answer is \emph{no}. $S$ deletes \emph{Lang was born
in Oehringen, W\"urttemberg, Germany.} and $C$ deletes \emph{Lang lead the
expedition, until the outbreak of World War I in 1914.} from the same
paragraph, with the same number of tokens. The reader answers \emph{No} with
$O$, $S$ and $C$, each a single content token ended by the native stop
token. SELD is $3.35\times10^{-3}$ for $S$ and $0.91\times10^{-3}$ for $C$;
JS divergence is $2.1\times10^{-7}$ and $9.8\times10^{-4}$; the
answer-likelihood drop is $5.2\times10^{-6}$ and $3.7\times10^{-2}$; KL
divergence at temperature 16 and $D$ also rank $S$ first, while KL at
temperature 1, the entropy change and self-judgment do not.

The case was selected for illustration from a fully enumerated pool and is
not a representative sample. Among pairs with unchanged answers, 535
reader--reference pairs are ranked correctly by SELD but not by JS
divergence, and 528 are ranked by SELD no higher than their control; the latter pool is
the honest counterpart, and the pass rates in the main text summarize both.
In Test 2, 230 question--reader slots with identical answers under $O$, $A$,
$B$ and $C$ are ranked correctly by SELD but not by JS divergence, and 244
are ranked by SELD no higher than the control. For example, for \emph{Are 11-11-11
(Film) and Bloodsport (Film) from the same country?}, Olmo-3-7B gives the same
answer under all four contexts (official answer \emph{yes}), and SELD, LUMINA-ECU, $D$ and the
answer-likelihood drop rank the support-removing swap first, whereas JS
divergence does not.

\FloatBarrier
\partheading{Part II: Mechanism}

\section{Mechanism: Temperature and Proposition~\ref*{prop:unsaid}}
\label{app:temperature}
This is a post-hoc analysis of the complete strict Hotpot deletion cohort,
not a new independent confirmation. Each question retains all selected
support references; references are averaged within a question, and
questions have equal weight. The inference set has 457 questions, 474
references, and 391 source components. The fixed families contain 99 ranking,
75 LL, and 75 F1 contrasts, with 100,000 shared component-bootstrap draws.
No new model forward or answer generation is needed.

\paragraph{Fixed finite-vocabulary expansion.}
The high-temperature connection to centered logit matching is classical
\citep{hinton2015distilling}; we use the following expansion to interpret
the empirical scan, not as a new general compression guarantee.

Let $a,b\in\mathbb R^V$ be fixed vectors with their coordinate means removed.
Set $p_T=\operatorname{softmax}(a/T)$ and
$q_T=\operatorname{softmax}(b/T)$. Expanding log-sum-exp and the probabilities
around the uniform distribution gives
\begin{align}
 \KL(p_T\Vert q_T)&=\frac{\|a-b\|_2^2}{2VT^2}+O(T^{-3}),\\
 \JS(p_T,q_T)&=\frac{\|a-b\|_2^2}{8VT^2}+O(T^{-3}).
\end{align}
For example, $\log\sum_i e^{a_i/T}=\log V+
\|a\|_2^2/(2VT^2)+O(T^{-3})$ and
$p_{T,i}=V^{-1}+a_i/(VT)+O(T^{-2})$; substitution into
$\sum_i p_{T,i}\log(p_{T,i}/q_{T,i})$ yields the KL coefficient.
The JS coefficient follows by expanding around the mean of the two
distributions. Centered logits equal centered log-probabilities exactly,
because log-softmax subtracts a coordinate-constant scalar. The squared
distance decomposes into $(\|a\|_2-\|b\|_2)^2+
2\|a\|_2\|b\|_2\CL(a,b)$, so convergence to squared distance does not
generally imply convergence to cosine ranking.

\paragraph{Proof of Proposition~\ref*{prop:unsaid}.}
(i) Adding a constant $c$ to every coordinate of a vector leaves it
unchanged after centering, and multiplying a centered vector by $\lambda>0$
leaves its direction unchanged, so the cosine of the two centered vectors,
and hence SELD, is invariant to both. A softmax temperature $T$ divides
the logits by $T$, and log-softmax subtracts a coordinate-constant
log-normalizer, so the centered log-probabilities at any temperature are a
positive multiple of the centered logits. The cosine of two centered
vectors is by definition their Pearson correlation. (ii) Multiplying the
JS expansion above by $8VT^2$ gives $\|a-b\|_2^2+O(T^{-1})$, and
$\|a-b\|_2^2=\|a\|_2^2+\|b\|_2^2-2a^\top b=(\|a\|_2-\|b\|_2)^2+
2\|a\|_2\|b\|_2\,(1-\cos(a,b))$. \qed

\paragraph{The prediction confirmed.}
Proposition~\ref*{prop:unsaid}(ii) predicts that tempered divergences
move toward the pattern term that SELD measures as the temperature grows.
The complete scan on the development deletion cohort confirms this
(\cref{tab:temp-scan}): JS and KL divergence rise from 78--83\% at $T=1$ to
within 0.4 points of SELD at $T=16$, and the centered squared distance
matches as well. This is the theory at work rather than a competing method:
tempered divergences reach SELD only after a temperature has been chosen,
whereas SELD is their temperature-free limit and needs no tuning. The
choice is not innocuous, because raising the temperature of JS divergence
also lowers its ability to detect a changed answer in four of the eight
development combinations (main paper, Section~\ref*{sec:why}); SELD fixes
the trade-off by construction. The confirmatory comparisons of the main
paper use JS divergence at the standard temperature.

\begin{figure*}[tbp]
\centering\includegraphics[width=\textwidth]{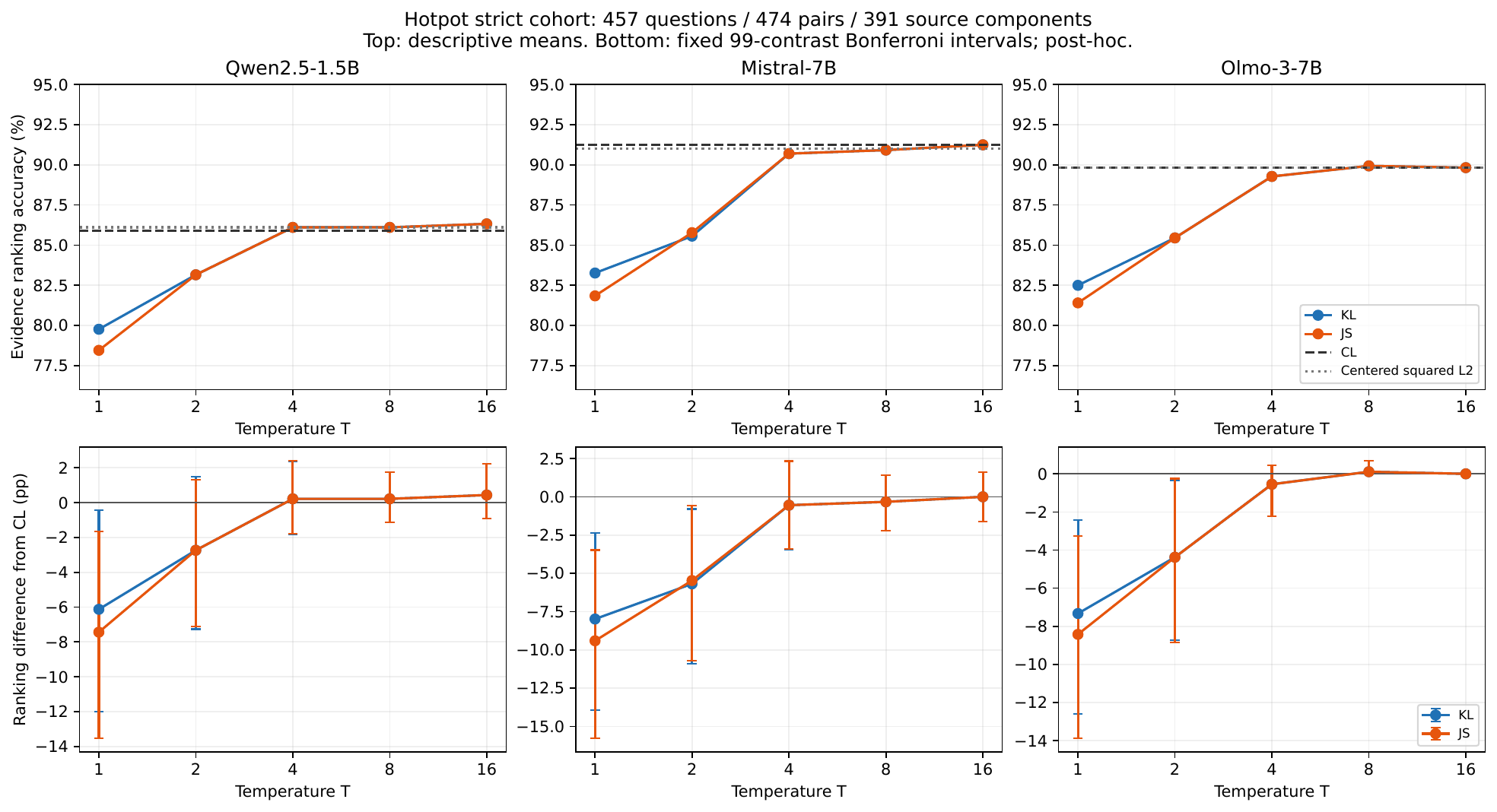}
\caption{Temperature ranking profiles and family-adjusted contrast
intervals from the complete post-hoc analysis. SELD and centered squared
distance remain separate references.}\label{fig:temp-mech}
\end{figure*}

\begin{table*}[tbp]
\centering\small
\begin{tabular}{llrrrrrrr}
\toprule
Reader & Score & $T=1$ & $T=2$ & $T=4$ & $T=8$ & $T=16$ & SELD & Squared $L_2$\\
\midrule
Qwen2.5-1.5B & KL & 79.8 & 83.2 & 86.1 & 86.1 & 86.3 & 85.9 & 86.1\\
Qwen2.5-1.5B & JS & 78.4 & 83.2 & 86.1 & 86.1 & 86.3 & 85.9 & 86.1\\
Mistral-7B & KL & 83.3 & 85.6 & 90.7 & 90.9 & 91.2 & 91.2 & 91.0\\
Mistral-7B & JS & 81.8 & 85.8 & 90.7 & 90.9 & 91.2 & 91.2 & 91.0\\
Olmo-3-7B & KL & 82.5 & 85.4 & 89.3 & 89.9 & 89.8 & 89.8 & 89.8\\
Olmo-3-7B & JS & 81.4 & 85.4 & 89.3 & 89.9 & 89.8 & 89.8 & 89.8\\
\bottomrule
\end{tabular}
\caption{Complete temperature scan on the development deletion cohort:
question-weighted evidence-ranking accuracy (\%). As predicted by
Proposition~\ref*{prop:unsaid}(ii), tempered divergences approach SELD.
Proximity is descriptive, not an equivalence test.}\label{tab:temp-scan}
\end{table*}
Only three of the 24 adjacent-temperature ranking comparisons have positive
family-adjusted intervals, all for Mistral-7B. No F1 contrast among the 75
fixed comparisons has an adjusted interval excluding zero; one LL contrast
(Qwen2.5-1.5B head-10 JS minus SELD) is negative. The mechanism evidence therefore
concerns diagnosis, not an established general improvement in answer utility.
\Cref{fig:temp-mech} plots the temperature profiles.

\section{Vocabulary Ablation}
\label{app:vocab}
For the ablation in \cref*{fig:vocab} of the main paper, the tokens are ranked by their
probability under $O$ (ties by token id). \emph{Top $m$} keeps only those $m$
coordinates of both vectors; \emph{all but top $m$} removes them. The selected
coordinates are re-centered before the angle is taken. \Cref{tab:vocab}
gives the per-cell values among unchanged answers of Test 1. The ordering
top~10 $<$ top~100 $<$ top~1000 $\le$ full holds in every cell, and removing the
top 100 or top 1,000 tokens changes the pass rate by at most 1.1 points. The
development data show the same pattern: the top-10 angle is below the full
angle in all eight cells, and removing the top 100 or 1,000 tokens gives
values close to the full angle. These are descriptive analyses; the
per-cell intervals are not corrected for a new family.

\begin{table}[htbp]
\centering\small
\setlength{\tabcolsep}{2.2pt}
\resizebox{\columnwidth}{!}{%
\begin{tabular}{llrrrrrr}
\toprule
Data & Reader & top 10 & top 100 & top 1k & full & $-$top 100 & $-$top 1k\\
\midrule
Hotpot & Qwen2.5-1.5B & 74.0 & 77.8 & 79.3 & 80.3 & 80.3 & 80.3\\
Hotpot & Mistral-7B & 77.3 & 82.5 & 84.6 & 86.0 & 86.0 & 85.7\\
Hotpot & Olmo-3-7B & 79.6 & 81.3 & 82.6 & 82.8 & 82.8 & 82.8\\
Hotpot & Qwen2.5-14B & 75.8 & 82.2 & 84.1 & 85.7 & 85.7 & 85.9\\
2Wiki & Qwen2.5-1.5B & 71.9 & 77.6 & 78.7 & 79.1 & 79.1 & 79.1\\
2Wiki & Mistral-7B & 70.4 & 76.3 & 79.8 & 79.8 & 79.8 & 79.8\\
2Wiki & Olmo-3-7B & 73.2 & 79.4 & 80.7 & 82.1 & 82.1 & 82.1\\
2Wiki & Qwen2.5-14B & 63.6 & 72.2 & 78.0 & 82.6 & 82.6 & 83.6\\
\midrule
Mean & & 73.2 & 78.6 & 81.0 & 82.3 & 82.3 & 82.4\\
\bottomrule
\end{tabular}}
\caption{Pass rate (\%) of the angle on vocabulary subsets, Test 1,
unchanged answers (post hoc). JS divergence and LUMINA-ECU average 68.9\%
and 65.4\% over the same cells.}
\label{tab:vocab}
\end{table}

\section{Hidden-State Angle in Test 1}
\label{app:hiddenD}
For all 4,459 pairs of Test 1, the cosine distance of the final normalized
hidden state ($D$) is available. Among pairs with unchanged answers (3,057
references), the reader-averaged pass rates are 82.7\% for SELD, 82.5\% for
$D$ and 70.3\% for JS divergence. SELD minus $D$ is +0.23 points
[$-0.44$, 0.92] (post hoc, corrected for 5, 100,000 draws), and $-0.08$ when
the answer changes. We therefore claim neither that SELD is better than
$D$ nor that they are equivalent. Since the logits are a linear map of the
normalized hidden state, both read the same reader state; SELD has the
advantage that it needs only the output distribution.

\partheading{Part III: Scope and Practical Use}

\section{Natural Compressor Outputs: Scope and Practical Use}
\label{app:natural}
The paired design of the main paper compares two edits of the same
context with exactly equal length, which isolates evidence from everything
else. Natural outputs offer no such control: outputs differ in length and
in what they keep, and support labels are only proxies for evidence. These
analyses therefore map how SELD behaves outside the paired setting and how
it is best deployed, rather than testing the paired claim.

\paragraph{Data.}
LLMLingua-2 at keep rates 25, 50 and 75\% and Provence (official reranker,
threshold 0.5) compress the full contexts of 4,831 questions (HotpotQA
1,742, 2Wiki 672, MuSiQue 2,417), giving 77,296 reader--output slots for the
four readers. Provence sees the question; LLMLingua-2 does not.

\paragraph{Clear-cut labels.}
A fixed analysis labelled an output with an unchanged answer as keeping
evidence if at least 90\% of the annotated support text survived and as
losing it if at most 10\% survived, and required 20 outputs per class per
cell. Only 6 of 48 reader--dataset--method cells qualified, all from
Provence; among outputs with unchanged answers, word-level pruning almost
never keeps at least 90\% of the support. In all six cells SELD is above JS divergence: its AUROC is 64.1, 91.6, 66.8, 61.7, 64.5 and
93.6\% against 60.6, 73.1, 58.1, 55.0, 56.4 and 64.7\% for JS divergence
(uncorrected pointwise intervals; the fixed pooled test is undefined).
The same self-judgment prompt answered by an external judge
(Qwen2.5-32B-Instruct) was also scored on these outputs: across the 23
estimable comparisons, all from Provence, SELD minus the judge is above zero
in 4, below zero in 8 and undetermined in 11 (post hoc, corrected for 384).
A larger external judge thus remains a strong option when its cost is
acceptable and the labels are clear-cut.

\paragraph{Combining with answer agreement.}
On the development data we mapped SELD and answer agreement to percentile
ranks calibrated on one half of the source components and evaluated on the
other. The alarm target is positive when the answer changed, or when it did
not change although a support sentence was deleted, and negative for an
unchanged answer after a control deletion. The mean of the two percentile
ranks improves the eight-cell AUROC over SELD by 2.83 [2.19, 3.52] and
over answer agreement by 16.23 [13.97, 18.57] points (corrected for the 7
comparisons of this family); the maximum of the two ranks improves on
answer agreement but by less than the preset 2 points on SELD. Two
further candidates that subtract a closed-book reference from SELD did
not improve the exam (0.54 [$-0.26$, 1.48] and $-13.2$ points).

\paragraph{Graded labels across outputs (post hoc).}
We then used the minimum retained fraction of the support spans as a
continuous label and, within each cell, scored all pairs of outputs with
unchanged answers whose overall retention differs by at most 5 points: a
pair counts as concordant when the output that kept less support has the
higher score. Over the 36 LLMLingua-2 cells (Provence had too few distinct
labels), SELD minus JS divergence is $-0.02$ [$-0.93$, $0.88$] and SELD
minus LUMINA-ECU $-0.07$ [$-1.03$, $0.88$] points (corrected for 2); per cell,
the three concordances lie between 44.4 and 55.8\%. Six sensitivity variants
(mean instead of minimum retention; 2- and 10-point length windows) give
intervals that all include zero.
Ranking different outputs by a graded label mixes evidence loss with how
much change each output tolerates, which the paired design removes; no
score resolves it here.

\paragraph{Choosing among outputs.}
For each question and reader we picked one of the four outputs by the
smallest score. Mean F1 over the four readers is 50.8 for SELD, 51.0 for
JS divergence and 50.0 for LUMINA-ECU, against 43.3 for a random choice,
47.1 for always using the 50\% LLMLingua-2 output and 66.4 for an oracle.
The candidates differ in length, so these gains are not length-controlled.

\paragraph{KV-cache compression.}
We attempted SnapKV \citep{li2024snapkv} and KVzip \citep{kim2025kvzip} for
all four readers. For six reader--method pairs, the implementations did not
reproduce the uncompressed reader within our numerical tolerance even
without compression (for SnapKV, the projections of the first layer already
differed when the computation was split); Olmo-3-7B is not supported by either
implementation. We therefore report no KV results.

\section{Real Compressors: Complete Strong Baselines}
\label{app:real}
Candidates are an abstractive prose summary, an abstractive fact list, an
extract of leading source units and LLMLingua-2 at requested budgets of
256, 512, 1,024 and 2,048 tokens, for MultiFieldQA-en (150), HotpotQA
(200), Qasper (200) and 2WikiMQA (200), with readers Qwen2.5-1.5B and
Mistral-7B. Selection happens within each budget and is averaged over the
four budgets. Scores were recomputed with full-vocabulary heads; generated
answers and F1 labels are those of the original runs.

\Cref{tab:real-full} lists all five scores on the LongBench candidates.
\begin{table}[htbp]
\centering\small
\setlength{\tabcolsep}{2.2pt}
\begin{tabular}{lrrrr}
\toprule
& \multicolumn{2}{c}{Qwen2.5-1.5B} & \multicolumn{2}{c}{Mistral-7B}\\
\cmidrule(lr){2-3}\cmidrule(lr){4-5}
Score & Sel.\ F1 & AUROC & Sel.\ F1 & AUROC\\
\midrule
$D$ & 32.8 & 70.9 & 37.6 & 79.3\\
SELD & 32.9 & 71.8 & 37.4 & 78.6\\
JS & 32.7 & 81.2 & 35.6 & 83.2\\
Answer agreement & 34.3 & 97.8 & 37.4 & 96.7\\
Own-answer drop & 33.2 & 81.4 & 36.1 & 83.9\\
\midrule
Random / oracle & \multicolumn{2}{c}{25.6 / 46.9} & \multicolumn{2}{c}{26.1 / 46.8}\\
\bottomrule
\end{tabular}
\caption{Selected F1 and answer-loss AUROC on candidates where all five
scores are valid. Selection gains over the own-answer drop: Mistral-7B SELD
$+1.3$ [0.1, 2.6], Qwen2.5-1.5B SELD $-0.3$ [$-1.6$, 1.1] (pointwise).}
\label{tab:real-full}
\end{table}

\paragraph{Gold-answer likelihood.}
A recomputation with full-vocabulary heads supports a
post-hoc family of 180 contrasts. Selecting by SELD rather than JS raises gold
log-likelihood by 0.196 nats/token [0.065, 0.339] for Mistral-7B; for Qwen2.5-1.5B the
gain (0.049) is not established. For Mistral-7B, SELD is a better alarm than JS
when the alarm target is a gold-likelihood drop of at least 1 nat/token
($+3.04$ AUROC points [0.17, 6.04]); thresholds 0.5 and 2 do not establish a
difference.

\paragraph{Scale calibration.}
Dividing the angle of each candidate by the angle of a fixed zero-overlap single
sentence deletion from the same question lowered answer-loss AUROC by 4.3
($D$) and 5.0 (SELD) points for Qwen2.5-1.5B and by 8.5 and 7.8 for Mistral-7B (pointwise
intervals all below zero). The calibrated angles remain 12.7--15.0 points
below JS and the own-answer drop. Within a question the calibration leaves
rankings unchanged.

\section{Cost}
\label{app:cost}
We time each score on one GPU with Qwen2.5-1.5B on synthetic QA prompts of
150 and 512 tokens (median of ten runs). Per candidate, SELD takes 29--30\,ms
and JS divergence 35--36\,ms. The answer-likelihood drop takes 23--24\,ms when
the original answer is cached and 84--86\,ms when it must be generated,
against 51--53\,ms for SELD including the pass over the original context.
These are synthetic measurements, not pipeline latencies.

\FloatBarrier
\partheading{Part IV: Development History and Reproducibility}

\section{Development Cohorts}
\label{app:dev}
\begin{figure*}[tbp]
\centering
\includegraphics[width=\textwidth]{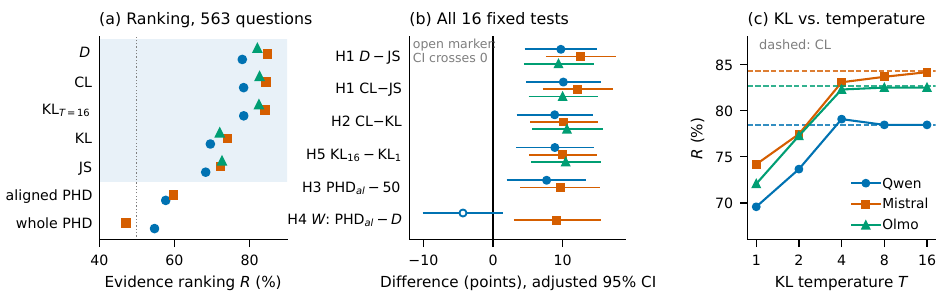}
\caption{Earlier 2Wiki development test (563 questions, 720 pairs,
456 components). (a) Ranking accuracy; (b) all 16 fixed tests with
Bonferroni 95\% intervals, including the unsupported Qwen2.5-1.5B order-response
test; (c) KL ranking over temperature, with SELD dashed.}
\label{fig:dev-2wiki}
\end{figure*}

The HotpotQA dev cohort (457 questions, 474 pairs, 391 source components
outside the largest component) and the 2Wiki dev cohort (563/720/456) were
used during exploration. For 2Wiki we fixed a 16-test family before reading
outputs (\cref{app:confirmation}); it established angle-over-JS advantages of
9.4--12.5 points for three readers (\cref{fig:dev-2wiki}). With four readers and the strong
baselines, the post-hoc 32-comparison family gives the values in
\cref{tab:p0}.

\begin{table}[htbp]
\centering\footnotesize
\setlength{\tabcolsep}{2.2pt}
\resizebox{\columnwidth}{!}{%
\begin{tabular}{llrrr}
\toprule
Data & Reader & SELD$-$JS & SELD$-\Own$ & SELD$-$judge\\
\midrule
Hotpot & Qwen2.5-1.5B & 8.0 & 22.0 & 6.7$^\circ$\\
Hotpot & Mistral-7B & 9.7 & 24.6 & 18.1\\
Hotpot & Olmo-3-7B & 7.3 & 22.9 & 7.9\\
Hotpot & Qwen2.5-14B & 13.5 & 22.4 & 8.9\\
2Wiki & Qwen2.5-1.5B & 10.0 & 21.5 & 10.8\\
2Wiki & Mistral-7B & 13.9 & 27.6 & 11.6\\
2Wiki & Olmo-3-7B & 9.4 & 27.5 & 8.2\\
2Wiki & Qwen2.5-14B & 15.5 & 28.4 & $-$0.0$^\circ$\\
\bottomrule
\end{tabular}}
\caption{Development cohorts: SELD minus JS, own-answer drop and
self-judgment (points, all references). Self-judgment is the reader model's
log-probability of \emph{No} minus \emph{Yes} for the prompt \emph{Determine
whether the passages alone contain enough information to uniquely answer the
question. Do not use outside knowledge or the plausibility of a guessed
answer. Answer exactly Yes or No.} All intervals of the 32-comparison
family exclude zero except those marked $^\circ$. SELD$-$answer agreement is
24.6--30.5.}
\label{tab:p0}
\end{table}

\section{Complete 2Wiki Confirmation}
\label{app:confirmation}
The full candidate frame has 1,621 questions and 2,072 references. Inference
uses the fixed 563 questions, 720 references and 456 components outside the
largest source component. The strict-length sensitivity frame contains 137
questions and 140 references outside that component; it is descriptive.
The 249-question transitive-isolation cohort is another descriptive
sensitivity, not a replacement inference set. Conservative length under the Qwen2.5 tokenizer
does not imply the same qualification under Mistral-7B or Olmo-3-7B tokenization.
No prompt or gold answer is truncated.

\begin{figure}[tbp]\centering
\includegraphics[width=\columnwidth]{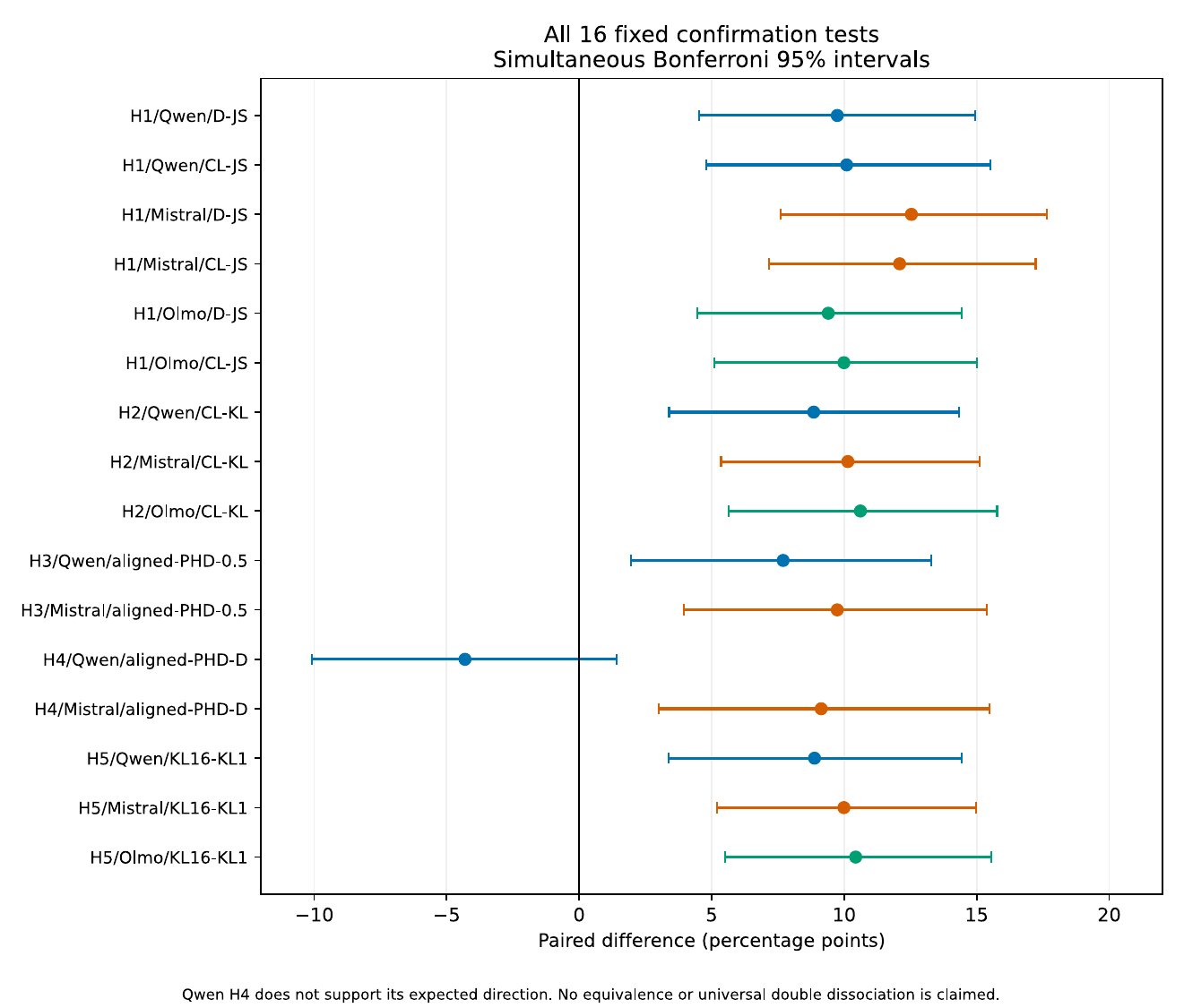}
\caption{Every fixed 2Wiki contrast, including the unsupported Qwen2.5-1.5B H4.}\label{fig:2wiki-all-tests}
\end{figure}

H1 comprises six angle-minus-JS comparisons, and passes only if at least two
readers pass both. H2 tests SELD minus KL in each reader; H3 and H4 test a
text-geometry score from the earliest phase of the project (persistent
homology dimension of the compressed text), which we do not pursue further
and report only for completeness; H5 tests KL at temperature 16 minus 1 in
three readers. All 16 tests share the fixed bootstrap draws and Bonferroni
95\% intervals. Qwen2.5-1.5B H4 fails and is retained. LL selection and all remaining
cohort descriptions are descriptive; this set adds no generated-answer F1
or independent natural-compressor utility confirmation.
\Cref{tab:2wiki-all-tests} and \cref{fig:2wiki-all-tests} list every fixed 2Wiki test.
\begin{table}[tbp]\centering\footnotesize
\setlength{\tabcolsep}{3pt}
\caption{All fixed 2Wiki tests with simultaneous 95\% intervals; differences are percentage points.}\label{tab:2wiki-all-tests}
\resizebox{\columnwidth}{!}{%
\begin{tabular}{lrrr}
\toprule Comparison & Difference (pp) & Lower & Upper\\\midrule
H1/Qwen2.5-1.5B/D-JS & +9.7 & +4.5 & +14.9\\
H1/Qwen2.5-1.5B/SELD-JS & +10.1 & +4.8 & +15.5\\
H1/Mistral-7B/D-JS & +12.5 & +7.6 & +17.7\\
H1/Mistral-7B/SELD-JS & +12.1 & +7.2 & +17.2\\
H1/Olmo-3-7B/D-JS & +9.4 & +4.4 & +14.4\\
H1/Olmo-3-7B/SELD-JS & +10.0 & +5.1 & +15.0\\
H2/Qwen2.5-1.5B/SELD-KL & +8.9 & +3.4 & +14.3\\
H2/Mistral-7B/SELD-KL & +10.1 & +5.4 & +15.1\\
H2/Olmo-3-7B/SELD-KL & +10.6 & +5.6 & +15.8\\
H3/Qwen2.5-1.5B/aligned-PHD-0.5 & +7.7 & +2.0 & +13.3\\
H3/Mistral-7B/aligned-PHD-0.5 & +9.7 & +3.9 & +15.4\\
H4/Qwen2.5-1.5B/aligned-PHD-D & -4.3 & -10.1 & +1.4\\
H4/Mistral-7B/aligned-PHD-D & +9.1 & +3.0 & +15.5\\
H5/Qwen2.5-1.5B/KL16-KL1 & +8.9 & +3.4 & +14.4\\
H5/Mistral-7B/KL16-KL1 & +10.0 & +5.2 & +15.0\\
H5/Olmo-3-7B/KL16-KL1 & +10.4 & +5.5 & +15.6\\
\bottomrule\end{tabular}}
\end{table}

\section{Near-Length Mistral-7B Extension}
\label{app:near-length}
This descriptive extension of \cref{app:temperature} to a larger near-length
cohort shows the same pattern: tempered divergences approach SELD, as
Proposition~\ref*{prop:unsaid} predicts. It uses all 14,486 fixed same-paragraph references
in 7,118 questions with saved full-vocabulary records. Outside the largest source
component it retains 4,745 questions, 9,511 references and 3,064 components.
No new forward, generation or inference test is performed. All 26 original
scores and complete gold LL are kept. Within each length stratum references
are averaged per question; strata can share questions and are not independent.
The conservative positive-deletion stratum contains 2,770 questions and
3,640 references; the equal positive-deletion stratum has 362 and 372.
Overall, all nine nonpositive-deletion references are retained and separately
described.

\begin{figure*}[tbp]\centering
\includegraphics[width=\textwidth]{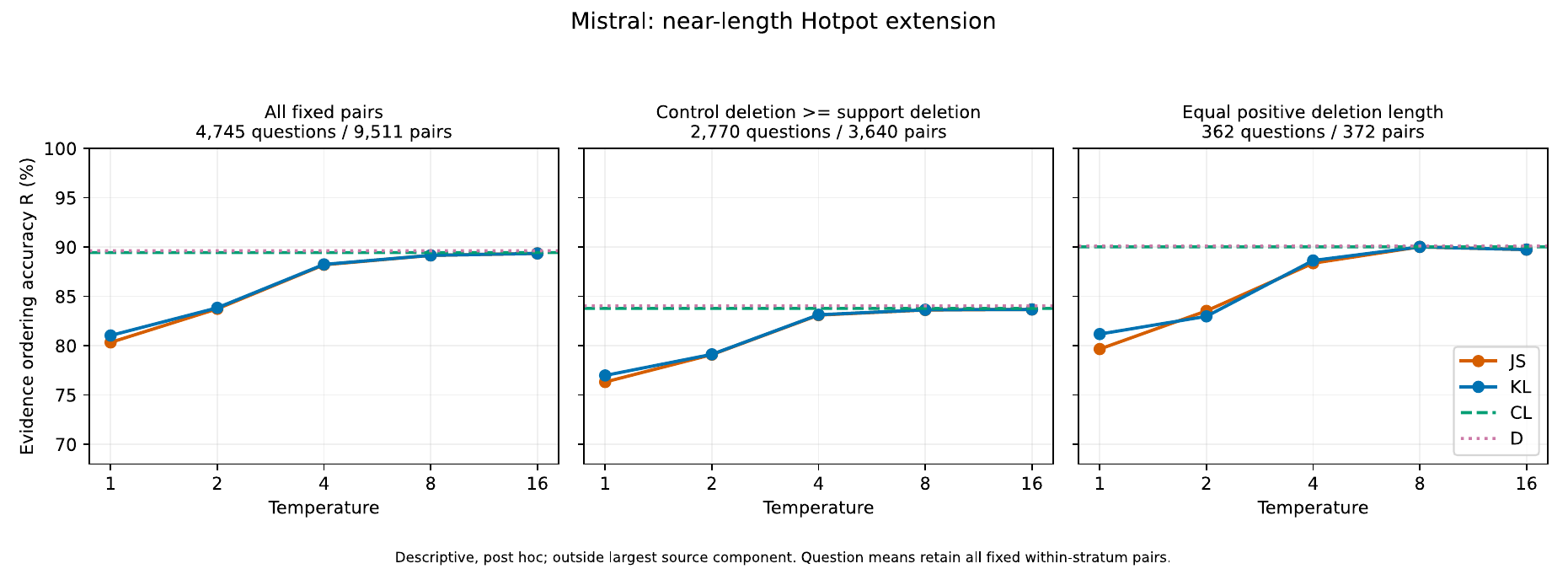}
\caption{Mistral-7B near-length temperature profiles, entirely descriptive.
The conservative stratum prevents greater support-token removal from
explaining the positive evidence ordering there. This does not establish
equivalence, strict monotonicity or answer-utility improvement.}\label{fig:near-mistral}
\end{figure*}

\begin{table}[tbp]\centering\footnotesize
\setlength{\tabcolsep}{4pt}
\caption{Mistral-7B near-length frame: evidence-ordering accuracy (\%) of each score over all fixed pairs and the conservative and equal-length strata (descriptive).}\label{tab:near-mistral}
\resizebox{\columnwidth}{!}{%
\begin{tabular}{lrrr}\toprule
Score & All fixed pairs & Conservative & Equal length\\\midrule
JS\_T1 & 80.3 & 76.3 & 79.7\\
KL\_T1 & 81.0 & 77.0 & 81.2\\
JS\_T16 & 89.3 & 83.7 & 89.7\\
KL\_T16 & 89.3 & 83.7 & 89.7\\
SELD & 89.4 & 83.8 & 90.0\\
D & 89.6 & 84.0 & 90.1\\
SELD\_tail1000 & 89.5 & 83.9 & 90.0\\
\bottomrule\end{tabular}}
\end{table}
\Cref{fig:near-mistral} plots these profiles.

\section{Near-Length Qwen2.5-1.5B Extension}
\label{app:near-length-qwen}
This descriptive extension repeats \cref{app:near-length} with Qwen2.5-1.5B, using
the saved full-vocabulary records of the Qwen2.5-1.5B development runs and retaining 5,552 questions and 8,224 fixed same-paragraph
references. Another 6,262 references lack a support-deletion route and are
reported as uncovered. Outside the
largest source component the frame has 3,697 questions, 5,469 references,
and 2,522 components. This differs from the Mistral-7B coverage above.
The conservative positive-deletion stratum has 1,778 questions and 2,072
references; exact positive matching has 458 questions and 475 references.
All six nonpositive-deletion references remain in the overall frame
(five outside the largest component). Within-stratum references are
averaged per question, retaining all fixed pairs. No new model forward, generation or inference test is
added.

\begin{table}[tbp]\centering\footnotesize
\setlength{\tabcolsep}{4pt}
\caption{Qwen2.5-1.5B near-length frame: evidence-ordering accuracy (\%) of each score over all fixed pairs and the conservative and equal-length strata (descriptive).}\label{tab:near-qwen}
\resizebox{\columnwidth}{!}{%
\begin{tabular}{lrrr}\toprule
Score & All fixed pairs & Conservative & Equal length\\\midrule
JS\_T1 & 80.7 & 76.4 & 78.5\\
KL\_T1 & 80.6 & 76.5 & 79.8\\
JS\_T16 & 87.4 & 81.4 & 86.4\\
KL\_T16 & 87.3 & 81.4 & 86.4\\
SELD & 87.4 & 81.4 & 85.9\\
D & 87.7 & 82.2 & 87.1\\
SELD\_tail1000 & 87.4 & 81.4 & 85.9\\
\bottomrule\end{tabular}}
\end{table}

\begin{figure*}[tbp]\centering
\includegraphics[width=\textwidth]{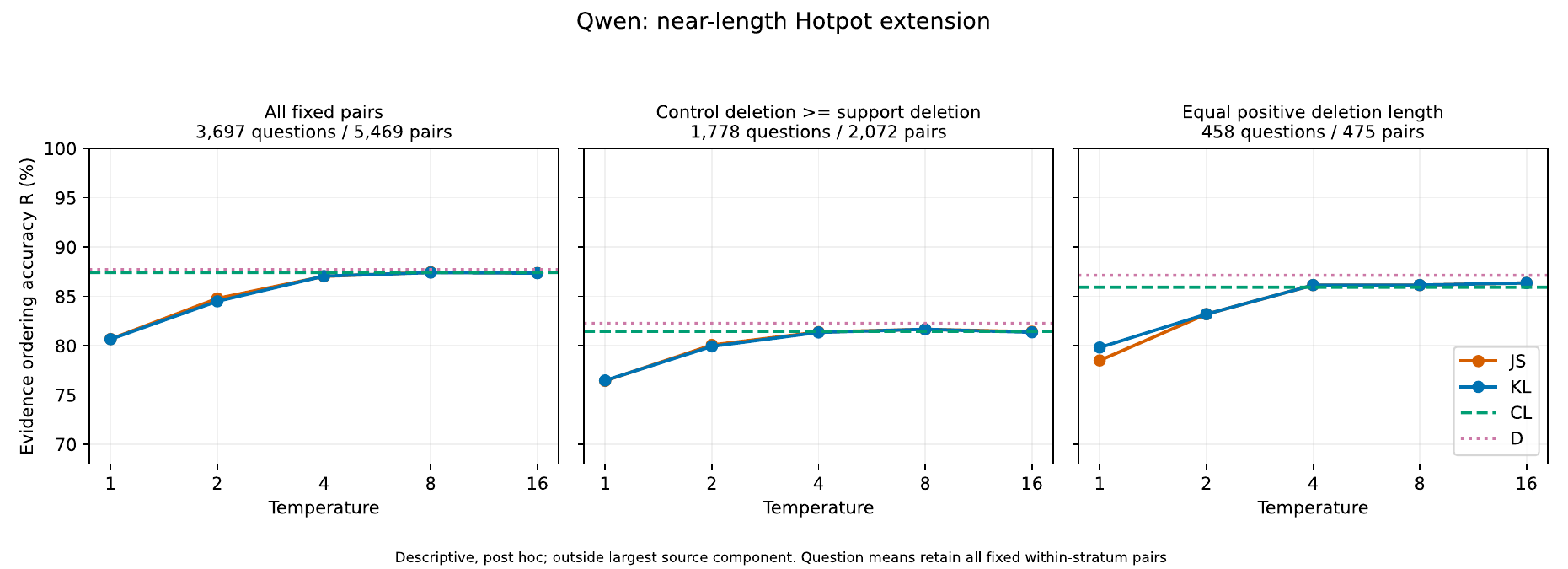}
\caption{Full descriptive Qwen2.5-1.5B temperature profiles, including the
nonmonotonic high-temperature segment and each fixed length stratum.
Coverage is distinct from the Mistral-7B extension.}\label{fig:near-qwen}
\end{figure*}

The conservative JS curve rises from 76.4\% at temperature 1 to 81.4\% at
16, against SELD at 81.4\% and D at 82.2\%. It declines slightly after
temperature 8, so the trend is not strictly monotone. Over all component-
excluded questions, selection by SELD versus JS at temperature 1 gives mean
gold LL of $-1.411628$ versus $-1.448471$ nats/token, a descriptive difference
of $+0.036843$ without a new confidence interval. Neither this difference
nor similar ranking values establish utility improvement or equivalence.
\Cref{fig:near-qwen} plots these profiles.

\section{Development Cohorts: Implementation Details}
\label{app:impl}
\paragraph{Prompts and decoding.}
Every reader uses its native tokenizer and chat template, BF16 weights and
greedy decoding with a 32-token answer cap (the official caps on LongBench).
First-token distributions are taken at the final prompt position, before any
answer token; the gold-answer likelihood is teacher-forced over every token
of each gold answer. Since $\log p_X = z_X - \mathrm{LSE}(z_X)\mathbf 1$,
centering log-probabilities equals centering logits, so SELD and JS/KL are
computed from the same vectors and differ only in how they compare them.

\paragraph{Deletion cohort.}
Control sentences exclude every annotated supporting sentence and are chosen
to match the token length of the deleted support sentence as closely as
possible. The strict frame has 741 questions and 769 pairs; source
components are defined by shared source titles.

\paragraph{Compression candidates.}
The extract keeps complete leading source units up to the budget, and
LLMLingua-2 is run at the requested budget. Empty outputs and overshoots are
retained.

\section{Assets and Reproducibility}
\label{app:assets}
\paragraph{Models and data.}
\Cref{tab:model-assets} lists the models we use. All are used frozen, without
training, fine-tuning or quantization, and only for research, consistent
with their terms. HotpotQA is released under CC BY-SA 4.0, MuSiQue and
Qasper under CC BY 4.0, and the 2WikiMultihopQA repository under Apache-2.0;
the LongBench code is MIT. For 2WikiMultihopQA and LongBench the
repositories do not specify a separate license for the underlying texts, and
we do not redistribute them.

\begin{table}[htbp]\centering\footnotesize
\setlength{\tabcolsep}{3pt}
\caption{Models used, all frozen; licenses as declared by the official
model cards.}\label{tab:model-assets}
\resizebox{\columnwidth}{!}{%
\begin{tabular}{llr}
\toprule Model & Use & License\\\midrule
Qwen2.5-1.5B/14B-Instruct & reader & Apache-2.0\\
Mistral-7B-Instruct-v0.3 & reader & Apache-2.0\\
Olmo-3-7B-Instruct & reader & Apache-2.0\\
LLMLingua-2 (XLM-RoBERTa-large) & compressor & MIT\\
bge-base-en-v1.5 & text-only baseline & MIT\\
bge-reranker-v2-m3 & text-only baseline & Apache-2.0\\
\bottomrule\end{tabular}}
\end{table}

\paragraph{Computing and reproducibility.}
Readers run in BF16 on RTX 3090 GPUs (24\,GB); the 14B reader is split
across two GPUs. All reported statistics were recomputed from the saved
scores by a second, independent implementation. Saved scores, edit
specifications and code will be released.

\paragraph{QA prompt.}
The following user message is filled and passed through each reader's
native chat template.
\begin{quote}\small
Answer the question based on the given passages. Only give me the answer and
do not output any other words.\par\medskip
The following are given passages.\par
\{context\}\par\medskip
Answer the question based on the given passages. Only give me the answer and
do not output any other words.\par\medskip
Question: \{input\}\par
Answer:
\end{quote}

\FloatBarrier

\end{document}